\documentclass[lettersize,journal]{IEEEtran}
\usepackage{booktabs}
\usepackage{amssymb}
\usepackage{makecell}
\usepackage{tabularx}
\usepackage{hhline}
\usepackage{multirow}
\usepackage{amsmath,amsfonts}
\usepackage{algorithmic}
\usepackage{algorithm}
\usepackage{array}
\usepackage[caption=false,font=normalsize,labelfont=sf,textfont=sf]{subfig}
\usepackage{textcomp}
\usepackage{stfloats}
\usepackage{url}
\usepackage{verbatim}
\usepackage{graphicx}
\usepackage{cite}
\usepackage{color}
\usepackage{xcolor}
\newcommand{\tcg}[1]{\textcolor{green}{\textbf{#1}}}
\newcommand{\tcr}[1]{\textcolor{red}{\textbf{#1}}}
\usepackage[colorlinks=true, allcolors=blue]{hyperref}
\usepackage{cleveref}
\newcommand{\nounderlineurl}[1]{{%
  \hypersetup{hidelinks}%
  \href{#1}{\nolinkurl{#1}}%
}}
\begin{document}
\title{FSG-Net: Frequency-Spatial Synergistic Gated Network for High-Resolution Remote Sensing Change Detection}
\author{Zhongxiang Xie, Shuangxi Miao, Yuhan Jiang, Zhewei Zhang, Jing Yao,~\IEEEmembership{Member,~IEEE}, Xuecao Li, \\Jianxi Huang,~\IEEEmembership{Senior Member,~IEEE}, and Pedram~Ghamisi,~\IEEEmembership{Senior Member,~IEEE}
\thanks{This work was supported by the National Key R\&D Program of China under Grant (2023YFB3907600). }
% (Corresponding author: Shuangxi Miao.)
\thanks{Z. Xie, S. Miao, Y. Jiang, Z. Zhang, and X. Li are with the College of Land Science and Technology, China Agricultural University, Beijing 100193, China (zx\_xie@cau.edu.cn; miaosx@cau.edu.cn; jiangyuhan@cau.edu.cn; zhewei@cau.edu.cn; xuecaoli@cau.edu.cn).}
\thanks{J. Yao is with the Aerospace Information Research Institute, Chinese Academy of Sciences, Beijing 100094, China (jasonyao92@gmail.com).}
\thanks{J. Huang is with the Faculty of Geosciences and Engineering, Southwest Jiaotong University, Chengdu 60031, China (jxhuang@swjtu.edu.cn).}
\thanks{P. Ghamisi is with Helmholtz-Zentrum Dresden-Rossendorf, Helmholtz Institute Freiberg for Resource Technology, 09599 Freiberg, Germany, and also with the Lancaster Environment Centre, Lancaster University, LA1 4YR Lancaster, U.K. (e-mail: p.ghamisi@gmail.com).}% <-this % stops a space
}
\maketitle
\begin{abstract}
Change detection from high-resolution remote sensing images lies as a cornerstone of Earth observation applications, yet its efficacy is often compromised by two critical challenges. 
First, false alarms are prevalent as models misinterpret radiometric variations from temporal shifts (e.g., illumination, season) as genuine changes. 
Second, a non-negligible semantic gap between deep abstract features and shallow detail-rich features tends to obstruct their effective fusion, culminating in poorly delineated boundaries.
To step further in addressing these issues, we propose the Frequency-Spatial Synergistic Gated Network (FSG-Net), a novel paradigm that aims to systematically disentangle semantic changes from nuisance variations. 
Specifically, FSG-Net first operates in the frequency domain, where a Discrepancy-Aware Wavelet Interaction Module (DAWIM) adaptively mitigates pseudo-changes by discerningly processing different frequency components. 
Subsequently, the refined features are enhanced in the spatial domain by a Synergistic Temporal-Spatial Attention Module (STSAM), which amplifies the saliency of genuine change regions. 
To finally bridge the semantic gap, a Lightweight Gated Fusion Unit (LGFU) leverages high-level semantics to selectively gate and integrate crucial details from shallow layers. 
Comprehensive experiments on the CDD, GZ-CD, and LEVIR-CD benchmarks validate the superiority of FSG-Net, establishing a new state-of-the-art with F1-scores of 94.16\%, 89.51\%, and 91.27\%, respectively. 
The code will be made available at \nounderlineurl{https://github.com/zxXie-Air/FSG-Net} after a possible publication.
\end{abstract}
\section{Introduction}
\IEEEPARstart{C}{hange} detection (CD) in remote sensing (RS) images is a vital technology for tracking geospatial changes~\cite{review1,wang2024hyperspectral}, with broad applications in urban planning~\cite{wen2019accurate}, land monitoring~\cite{pang2025special}, and disaster evaluation~\cite{disaster}. Still, the advent of vast sub-meter resolution datasets from sensors like WorldView, QuickBird, and Gaofen-2 has posed significant challenges to conventional methods like CVA~\cite{cva} and Random Forest~\cite{rf}, revealing their diminished efficacy~\cite{worldview}. This has paved the way for deep learning-based CD to become the dominant paradigm, owing to the inherent capability of architectures like Convolutional Neural Networks (CNNs) to automatically extract hierarchical features from large-scale, complex data~\cite{2024review}. Unfortunately, even these advanced deep learning models are not without their own set of critical obstacles. 

Technically, the objective of CD tasks is to identify semantic changes of interest at a pixel-wise level from two co-registered RS images acquired at different times. These differences in acquisition time inevitably lead to discrepancies in imaging conditions. Factors such as illumination intensity and seasonal turnover can introduce significant pseudo-changes--non-semantic variations like building shadows or phenological changes in vegetation (e.g., lawns before and after snow cover). Moreover, another challenge lies in the morphological ambiguity between objects of varying scales. For instance, the low-level textural patterns of a newly paved footpath might be nearly identical to those of a distant, large-scale construction site boundary, despite their clear conceptual disparity. 
This semantic gap between deep, high-level concepts and shallow, fine-grained details inevitably degrades the precision of boundary delineation. 
Therefore, to accurately identify changes of interest within complex scenarios, a trustworthy CD model is expected to embody the following two capabilities: 
\begin{itemize}
    \item \textbf{Pseudo-change suppression.} Distinguishing genuine changes of interest from a multitude of task-irrelevant interference.
    \item \textbf{Semantics-aware alignment and fusion.} Ensuring a coherent fusion of multi-level features, aligning abstract semantics with fine-grained spatial details to produce well-delineated boundaries.
\end{itemize}

Aiming at the aforementioned challenges, a growing body of research has emerged, exploring solutions from a variety of perspectives. For example, Tang et al.~\cite{31x31}, and Li~\cite{Dilated} have employed techniques like larger convolution kernels and dilated convolutions to incrementally expand the receptive fields of CNN models, thereby emphasizing the significance of environmental alterations~\cite{Poly}. Concurrently, the integration of attention mechanisms, such as self-attention (SelfAtt)~\cite{self-attention} and  cross-attention (CrossAtt)~\cite{Asymmetric}, has also become a prevalent strategy for amplifying the most relevant information~\cite{Coordinateattention}. Moreover, it is now widely recognized that facilitating feature interaction between bi-temporal representations, before the final difference computation, is paramount for robust change discrimination~\cite{ISPRSattention,DMINet,Featureinteraction}. For this purpose, Liu et al.~\cite{ISPRSattention}introduced a partial feature exchange mechanism in the channel and spatial dimensions to promote information sharing between different temporal images. Fang et al.~\cite{Featureinteraction} constructed a novel CD architecture named MetaChanger, which embeds a cascade of alternating interaction modules within the backbone feature extractor. Perceiving that purely spatial domain interaction faces a bottleneck in sufficiently disentangling authentic changes from background noise, the latest focus has gradually started shifting towards the frequency domain~\cite{sun2025mask,FTransDF-Net,frequency1}.

In signal processing, the transformation of an image from the spatial domain to the frequency domain leads to a distinct separation of the image's spectral content~\cite{46F}. Low-frequency components predominantly capture the smooth, global characteristics of the image, encapsulating variations in illumination and overarching background changes~\cite{48F}. Conversely, high-frequency components focus on the finer and more intricate details of the image, including regions of rapid intensity variation, such as edges, textures, and noise, which are critical for discerning detailed structures and boundary~\cite{38f}. Leveraging this principle, various techniques, including the Discrete Cosine Transform~\cite{frequency3}, Discrete Wavelet Transform (DWT)~\cite{WS-Net++}, and Fast Fourier Transform~\cite{frequency2}, have been utilized to decouple object identity from stylistic variations, thereby enhancing the model's robustness to irrelevant changes. Despite these advances, in tackling the conundrum of pseudo-changes, the preceding methods still fail to establish a collaborative mechanism between the frequency and spatial domains.

The effective fusion of multi-level features remains a challenge, as baseline approaches that rely on simple concatenation, addition, or bilinear pooling often fail to reconcile the representational gap between deep semantic features and shallow, fine-grained details~\cite{Featureinteraction,yao2023extended}. Consequently, it necessitates the development of a bridge mechanism to harmonize features across different hierarchical levels, ensuring proper alignment of high-level concepts and fine-grained details prior to their fusion. Recently, Chen et al.~\cite{frequency2} designed a U-fusion change perception module, which bidirectionally fuses change features across multiple scales to improve precise boundary delineation. Xu et al.~\cite{2024hybrid} employed a coarse-to-fine feature interaction module to progressively fuse multiscale features via a hierarchical strategy.  Furthermore, another work utilized flow field or deformable convolutions to spatially align multi-level features before their fusion~\cite{deformable2,deformable5}. While effective to some extent, such intricate, multi-scale interaction mechanisms often come at the cost of high computational complexity.

Answering this call, we introduce FSG-Net, a frequency–spatial synergistic framework that separates the change signal from the nuisance and reconciles feature asymmetry in a principled manner. The core idea behind FSG-Net lies in: \textbf{1)} adopting a synergistic frequency-spatial separation strategy to discern targets of interest, and \textbf{2)} leveraging deep semantics to refine multi-level feature details.

Concretely, we first introduce a Discrepancy-Aware Wavelet Interaction Module (DAWIM) that operates on multi-resolution wavelet sub-bands, where cross-temporal discrepancy cues drive adaptive sub-band interaction and re-weighting. This yields a learnable frequency mask that attenuates low-frequency radiometric shifts while preserving high-frequency structural evidence, and the modulated representation is re-projected to the spatial stream through a residual path. Building on these refined features, a Synergistic Temporal-Spatial Attention Module (STSAM) is proposed to couple the temporally enhanced CrossAtt (to exchange long-range contextual cues between the two times) with coordinate attention (to encode positional priors), thereby amplifying saliency in truly changed regions and stabilizing local structures. Finally, to bridge the semantic gap between deep and shallow features--a common cause of blurred change boundaries--we propose a Lightweight Gated Fusion Unit (LGFU). This module generates semantics-driven gates from deep features to selectively propagate fine-grained details from earlier layers, yielding crisp boundary delineation with minimal computational overhead. To summarize, the main contributions of this work are as follows,

\begin{enumerate}
    \item We propose the Frequency-Spatial Synergistic Gated Network (FSG-Net), a novel architecture that pioneers a frequency-spatial synergistic pipeline to systematically disentangle semantic changes from nuisance variations in remote sensing images.
    \item To suppress background interference arising from discrepancies in acquisition conditions, we introduce a wavelet interaction module, termed DAWIM, that applies tailored strategies to different sub-bands in the frequency domain, effectively mitigating false alarms from radiometric shifts.
    \item A novel spatial-temporal attention module STSAM is designed to synergistically amplify the saliency of genuine changes by simultaneously modeling global context and preserving fine-grained local details through coupling cross- and coordinate-attention.
    \item We devise a lightweight unit LGFU that bridges the deep-shallow semantic gap to sharp change boundaries. It features semantics-driven gates to selectively fuse features, yielding crisp boundary delineation with high computational efficiency.
\end{enumerate}

The remainder of this article is organized as follows. Section \ref{s2} reviews related work in deep learning-based change detection. Section \ref{s3} elaborates on the proposed FSG-Net, detailing the overall architecture and its core modules. Section \ref{s4} describes the experimental setup, presents a comprehensive comparison and analysis against state-of-the-art methods, and includes ablation studies to validate our design choices. Finally, Section \ref{s5} concludes the paper and discusses potential future work.

\section{Related Work}\label{s2}
\subsection{Deep Learning-Based Change Detection}
Driven by the powerful feature extraction and representation capabilities of CNNs, deep learning-based CD methods have burgeoned in recent years~\cite{if16.2}. Seminal architectures such as FCN, U-Net, and ResNet~\cite{ResNet}, when augmented with advanced modeling techniques like multi-scale strategies and Transformer, demonstrated impressive performance on various CD tasks\cite{Res-like1,Unet-like,MSCA}. From the perspective of information flow and processing paradigms, we broadly categorize existing CD methods into three classes: CNN-based, Transformer-based, and the currently prevalent hybrid architectures.

\begin{figure*}[t]
    \centering
    \includegraphics[width=\textwidth]{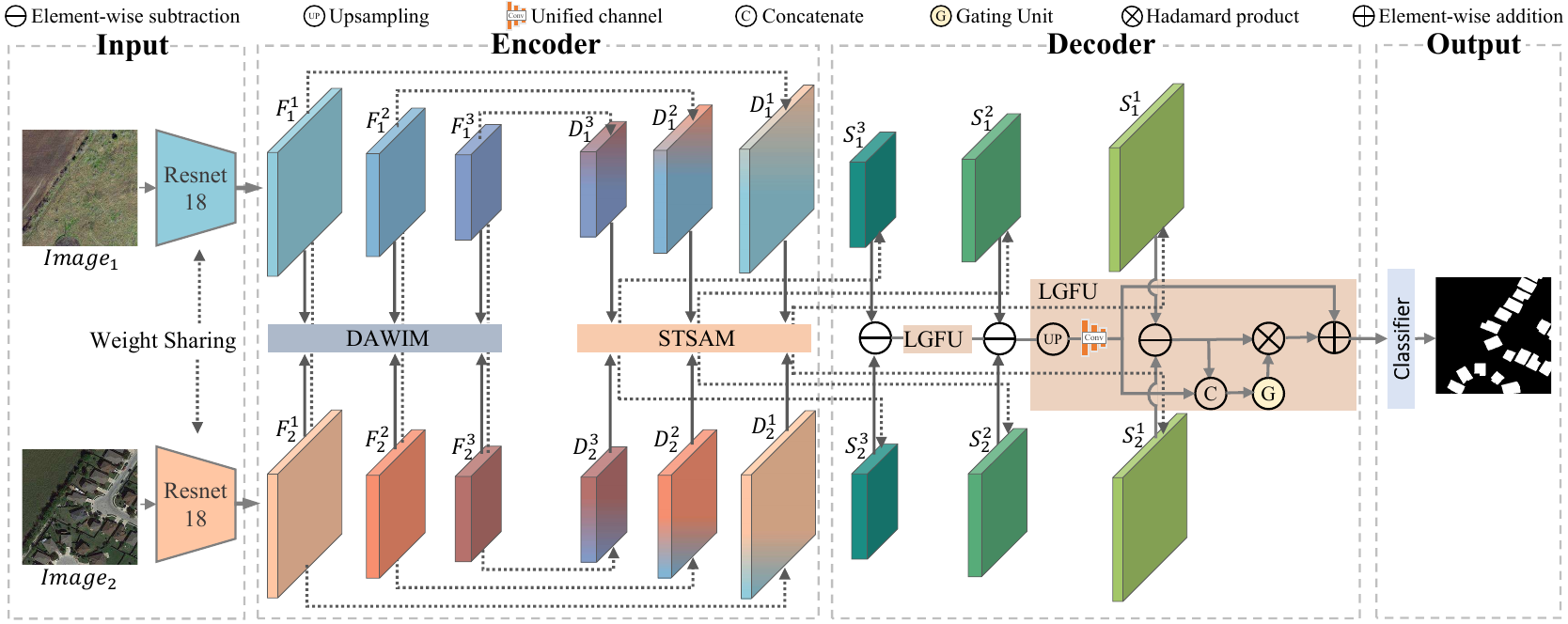}
    \caption{The overall architecture of the proposed FSG-Net, where $F_i^j$, $D_i^j$, and $S_i^j$ represent feature maps from the backbone, DAWIM, and STSAM, respectively. The subscript $i \in \{1, 2\}$ denotes the temporal phase, while the superscript $j \in \{1, 2, 3\}$ indicates the feature scale.}
    \label{FRAME}
\end{figure*}

Daudt et al.~\cite{FC-EE} first introduced a weight-sharing Siamese FCN framework to the CD task, presenting three variants that explored both early fusion (FC-EF) and feature-level fusion through concatenation (FC-Siam-Conc) and differencing (FC-Siam-Diff). Techniques such as large-kernel convolutions~\cite{31x31} and dilated convolutions~\cite{Dilated} are frequently employed to expand the effective receptive field, thereby enabling CNN models to capture richer contextual information. Simultaneously, various attention mechanisms, including spatial~\cite{SPATIAL} and channel attention~\cite{CHANNEL} have been widely adopted to enhance the saliency of genuine change regions. Despite the extensive exploitation of their potential, CNNs are inherently limited by their inability to model long-range dependencies, a shortcoming that leaves clear room for enhancement~\cite{review2}. 

The Transformer architecture, originally proposed for natural language processing, has exhibited remarkable success in modeling long-range dependencies through its SelfAtt~\cite{Attention}. Capitalizing on this strength, Dosovitskiy et al.~\cite{VIT} identified the inductive biases of CNNs (locality and translation equivariance) as a fundamental constraint on modeling long-range relationships, and pioneered the adaptation of the Transformer to the visual domain with their Vision Transformer (ViT). Subsequently, Bandara et al.~\cite{Changeformer} proposed a Siamese framework integrating a hierarchical Transformer encoder and an MLP decoder to efficiently model multi-scale context for CD task. Similarly, Zhang et al.~\cite{swinsunet} designed a pure transformer network, named SwinSUNet, with Siamese U-shaped structure, featuring a remarkable 32-block hierarchy. Nevertheless, pure Transformers are often data-hungry and computationally demanding, struggling to capture the fine-grained boundary details at which CNNs excel.

To combine the strengths of both frameworks, Chen et al.~\cite{BIT} pioneered a hybrid model, BIT, that utilizes a Transformer encoder to capture global context from CNN features via SelfAtt. Subsequently, hybrid architectures have increasingly emerged as the dominant approach in the field of CD~\cite{ConFormer-CD,CDNeXt}. Yang et al.~\cite{ConFormer-CD} proposed a hybrid model ConvFormer with parallel convolution and multihead SelfAtt enhances the adaptability to complex scenes. GCFormer~\cite{GCFormer} features a Multi-Receptive-Field Convolutional Attention mechanism, specifically designed to extract global context from multiple receptive fields. Informed by prior advancements, we strategically incorporated an enhanced CrossAtt into a CNN architecture, aiming to strike a 
balance between detection accuracy and computational efficiency.
\subsection{Bi-temporal Feature Interaction in Change Detection}
A fundamental characteristic that distinguishes CD from other dense prediction tasks is the necessity of modeling interactions between bi-temporal features. The importance of feature interaction, well-documented across both homogeneous and heterogeneous data, is especially pertinent to CD due to the domain gap caused by variations in imaging conditions~\cite{Featureinteraction1}. Numerous approaches have been proposed to address this challenge. Feng et al.~\cite{DMINet} devised an intertemporal joint-attention block to manipulate the global feature distribution of each input, thereby stimulating information coupling between intra-level representations. Fang et al.~\cite{Featureinteraction} advocated for early feature interaction via a co-attention mechanism, demonstrating the critical importance of interaction before fusion by treating features directly as attention maps. Zhao et al.~\cite{MFCD} designed the FMCD model to enhance detection in complex scenes by modeling feature context through a multi-level interaction module, yielding highly representative features. While these spatial interaction methods have shown promise, they still operate on features where semantic content and stylistic variations are entangled.

Recently, the focus of feature interaction research has begun to shift from the spatial domain towards an inclusion of the frequency domain. Xue et al.~\cite{frequency3} enhanced feature representations by applying the Discrete Cosine Transform and then using Global Average Pooling on each frequency component to model channel-wise correlations. Tang et al.~\cite{frequency1} proposed FDINet for fine-grained object CD, which utilizes Frequency Decoupling Interaction to boost the model's discriminative power for different change types. Chen et al.~\cite{frequency2} devised the FIMP model, which utilizes the Fourier transform to perform adaptive filtering of bi-temporal features in the frequency domain, thereby isolating and enhancing task-specific features for CD. Regrettably, existing methods have insufficiently probed the feasibility of establishing a collaborative mechanism between frequency-domain and spatial-domain feature interactions to fully unlock the potential of their combined strengths.

Combining the strengths of existing research, we seek to construct a new paradigm for frequency-spatial domain interaction. In this paradigm, frequency-domain interaction is used to suppress pseudo-changes, while spatial domain interaction serves to enhance genuine changes.

\section{Methodology}\label{s3}
\subsection{Method Overview}
The overall architecture of FSG-Net, illustrated in Fig.~\ref{FRAME}, is designed as a principled, multi-stage pipeline that systematically addresses the core challenges in CD. This process begins with the DAWIM, which purifies the feature representations by suppressing pseudo-changes in the frequency domain. Complementing this, the STSAM then enhances the saliency of authentic change regions in the spatial domain. Finally, the LGFU harmonizes the multi-level features to ensure precise boundary delineation. This cascaded refinement strategy guarantees that both task-irrelevant interference and the semantic gap are collectively mitigated.
\subsection{Discrepancy-Aware Wavelet Interaction Module (DAWIM)}
\begin{figure}[t!]
    \centering
    \includegraphics[width=0.9\columnwidth]{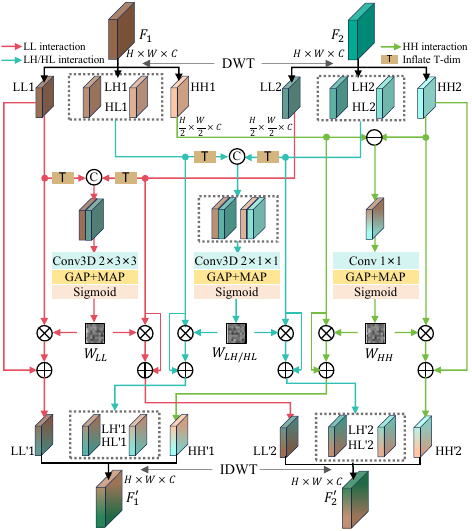}
    \caption{The operational workflow of the DAWIM.}
    \label{DAWIM}
\end{figure}

While the principle of frequency-domain interaction is promising, the key challenge lies in how to process the different frequency components to simultaneously enhance change signals while suppressing noise. The Wavelet Transform offers advantages such as time-frequency localization and multi-frequency subband analysis, making it particularly well-suited for CD tasks involving scale-sensitive targets~\cite{WS-Net++}. Accordingly, the DAWIM employs distinct processing strategies tailored to the specific characteristics of each frequency component. The operational workflow of this module is illustrated in Fig.~\ref{DAWIM}.

Specifically, a 2D Haar DWT is first applied to decompose the bi-temporal spatial feature maps, denoted as $F_i$ (where $i \in \{1, 2\}$), into four frequency-domain subcomponents, denoted as $\{LL_i, LH_i, HL_i, HH_i\}$. This decomposition is achieved by applying a set of predefined wavelet filters, followed by downsampling ($\downarrow$), as formulated below:
\begin{align}
    LL_i &= (F_i * f_{\text{low}} * f_{\text{low}}^{T}) \downarrow \label{eq:f_ll}, \\
    LH_i &= (F_i * f_{\text{low}} * f_{\text{high}}^{T}) \downarrow \label{eq:f_lh}, \\
    HL_i &= (F_i * f_{\text{high}} * f_{\text{low}}^{T}) \downarrow \label{eq:f_hl}, \\
    HH_i &= (F_i * f_{\text{high}} * f_{\text{high}}^{T}) \downarrow, \label{eq:f_hh}
\end{align}
where $f_{\text{low}} = [1/\sqrt{2}, 1/\sqrt{2}]$ and $f_{\text{high}} = [1/\sqrt{2}, -1/\sqrt{2}]$ are the low-pass and high-pass filters, respectively. The superscript $T$ indicates the transpose operation for applying the filters in the vertical direction. Subsequently, each frequency component undergoes an appropriate processing strategy based on its characteristics to maximize the benefits of frequency feature interaction.

The low-frequency components $LL_i$ primarily represent large-scale spatial patterns and gradual radiometric variations, such as land cover distribution and global illumination trends~\cite{47F}. Meanwhile, they exhibit relatively stable changes in the time domain, primarily reflecting gradual background variations~\cite{48F}. To capture these slow temporal dynamics effectively, a 3D convolution is applied to extract and then enhance the temporal features embedded within these frequency bands. Specifically, a temporal dimension is first inflated to $LL_i \in \mathbb{R}^{C\times H \times W}$, forming $\hat{LL}_i \in \mathbb{R} ^{C \times 1 \times H \times W}$. Then, $\hat{LL}_1$ and $\hat{LL}_2$ are concatenated along the temporal dimension to form $LL \in \mathbb{R} ^{C \times 2 \times H \times W}$ and a 3D convolution is applied to enhance the temporal relationship modeling between the bi-temporal features. The kernel size is set to $2 \times 3\times 3$, to perform feature fusion along the temporal dimension, while capturing spatial context from a $3 \times 3$ neighborhood. Finally, a reshaping operation is performed on the features $\hat{LL} \in \mathbb{R} ^{C \times 1 \times H \times W}$ obtained from the 3D convolution to embed the temporal dynamics into the channels, resulting in low-frequency interaction features $\widetilde {LL} \in \mathbb{R} ^{(C \times 1) \times H \times W}$. The entire operation can be formulated as:
\begin{equation}
    \label{eq:ll_reshape} 
    \widetilde{LL} = \text{Reshape}(\text{Conv3D}_{2 \times 3 \times 3}(\text{Concat}[\hat{LL}_1, \hat{LL}_2])),
\end{equation} 
where $\text{Concat}$ represents concatenation along the temporal dimension.

Compared to the $LL_i$, the mid-frequency components $(LH_i/ HL_i)$ encapsulate more localized temporal variations, particularly abrupt structural changes in the horizontal and vertical directions~\cite{wavelet1}. We therefore adopt an interaction method similar to that used for components $LL_i$, using 3D convolution to form $\widetilde{LH}$ and $\widetilde{HL}$, but set the kernel size to $2 \times 1 \times 1$ to fuse features along the temporal dimension without compromising the spatial structure.
\begin{align}
   \widetilde{LH} &= \text{Reshape}(\text{Conv3D}_{2 \times 1 \times 1}(\text{Concat}[\hat{LH}_1, \hat{LH}_2]))\label{eq:widetilde{LH}}, \\
    \widetilde{HL} &= \text{Reshape}(\text{Conv3D}_{2 \times 1 \times 1}(\text{Concat}[\hat{HL}_1, \hat{HL}_2]))\label{eq:widetilde{HL}}.
\end{align}

The high-frequency component $HH_i$, typically associated with edges, contours, and fine-grained structural details, is processed using element-wise subtraction to obtain the difference feature~\cite{WS-Net++}, formulated as follows:\begin{equation}
\label{eq:hh_diff}
\widetilde{HH} = \delta(\text{BN}(\text{Conv}_{1 \times 1}(|HH_1 - HH_2|))).
\end{equation}

After obtaining these frequency interaction features, we introduce an adaptive weighting mechanism based on modified Squeeze-and-Excitation Networks (SENet)~\cite{SENet}. This approach dynamically enhances the saliency of change regions while mitigating the impact of irrelevant background noise. Taking $\widetilde{LL}$ as an example, both max pooling $(MAP)$ and global average pooling $(GAP)$ are applied to the $\widetilde{LL}$ simultaneously—the former highlights the most significant feature responses, while the latter ensures that the model does not focus solely on extreme variations but instead learns more robust features. Then, a global feature vector, obtained by concatenating the max-pooled and average-pooled features, is used to compute the channel-adaptive weight $W_{\widetilde{LL}}$. Finally, the computed weight is applied to the original $LL_i$ component via the Hadamard product, followed by a residual connection. This operation aims to leverage change-aware features, derived from bi-temporal interaction,  to guide and refine the original, mono-temporal representations. The formulation is described as follows:
\begin{align}
    W_{\widetilde{LL}} &= \sigma(W_2\delta(W_1[\text{GAP}(\widetilde{LL}), \text{MAP}(\widetilde{LL})])) \label{eq:W_LL}, \\
    LL'_i &= W_{\widetilde{LL}} \odot LL_i + LL_i \label{eq:LL_prime}.
\end{align}

Here, $\sigma$ is the Sigmoid activation function, $\delta$ is the ReLU activation unit, and $W_1$, $W_2$ are the fully connected layer weights, where $W_1$ reduces dimensionality and $W_2$ restores it. The symbol $\odot$ denotes the Hadamard product.

Following the interaction between different frequency components, the feature change maps are reconstructed to the spatial domain through the application of the inverse discrete wavelet transform (IDWT). It can be formulated as:
\begin{equation}
    F'_i = \text{IDWT}(LL'_i, LH'_i, HL'_i, HH'_i) \label{eq:idwt}.
\end{equation}

\subsection{Synergistic Temporal-Spatial Attention Module (STSAM)}

Following the frequency-domain feature interaction, noise interference induced by acquisition conditions is effectively attenuated. Our subsequent objective is to leverage spatial feature interaction to establish an information conduction pathway between the bi-temporal features, thereby mutually steering the allocation of attention to regions with genuine changes. To accomplish this, the STSAM is constructed by integrating a CrossAtt augmented with temporal embeddings and a coordinate-attention (CoordAtt)~\cite{Coordinateattention}. Compared to the SelfAtt, the CrossAtt equally processes the features of both temporal instances, avoiding bias towards a single image while simultaneously capturing temporal differences and global changes. Complementing this global temporal interaction, the CoordAtt is further employed to refine the local spatial structure, ensuring that fine-grained details, such as changes in building boundaries, are given adequate attention.

\begin{figure}[t!]
    \centering
    \includegraphics[width=1.0\columnwidth]{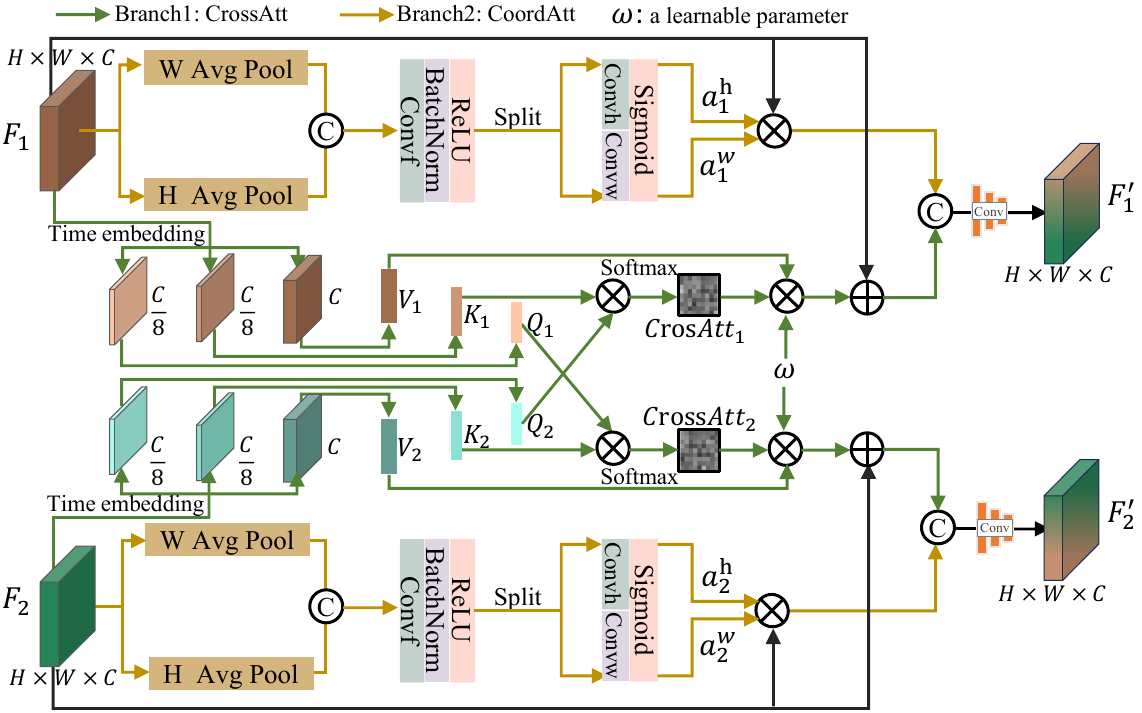}
    \caption{The internal mechanism of the STSAM.}
    \label{stsam_architecture}
\end{figure}

As shown in Fig.~\ref{stsam_architecture}, $F_i \in \mathbb{R}^{C \times H \times W}$ are fed into a parallel processing stage consisting of two branches: the CrossAtt (branch~1) and the CoordAtt (branch~2). In branch~1, learning temporal embeddings (inspired by position embeddings in Transformer)~\cite{Attention}, denoted as $T_i \in \mathbb{R}^{C \times 1 \times 1}$, are first introduced and then added to $F_i$ through Python's broadcasting mechanism, aiming to amplify the temporal discrepancies between inputs. Next, the enhanced features are projected into query ($Q_i$), key ($K_i$), and value ($V_i$) spaces using shared $1 \times 1$ convolutions. To maintain computational efficiency and in line with common practice, the channel dimensions of $Q$ and $K$ are compressed to 1/8 of the original, while the dimension of $V$ is preserved.

Subsequently, the $Q$ from the counterpart is used to perform similarity computation and weighting operations within its own $K-V$ space, ultimately yielding the final CrossAtt output. This output is then scaled by a learnable parameter $\omega$ and added back to the original feature $F_i$ via a residual connection. The process is described by the following formulas:
\begin{align}
    Q_i, K_i, V_i &= \text{Linear}(\text{Conv}_{1 \times 1}(F_i + T_i)), \\
    CrossAtt_i &= \text{Softmax}(Q_j \cdot K_i^T) \cdot V_i, \\
    F^{\text{branch1}}_i &= \omega \cdot CrossAtt_i + F_i,
\end{align}
where $i, j \in \{1, 2\}$ are temporal indices with $i \neq j$, and $\omega$ is a learnable scaling parameter initialized to 0 for training stability.

In branch~2, adaptive average pooling is first applied to $F_i\in \mathbb{R}^{C\times H \times W}$ along the height and width directions, resulting in two directional feature map tensors: $x_i^h \in \mathbb{R}^{C\times H\times 1}$ and $x_i^w\in \mathbb{R}^{C\times 1\times W}$ (after dimension transposition). Then, $x_i^h$ and $x_i^w$ are concatenated along the spatial dimension and processed through a 1×1 convolution to fuse the pooled information from both directions. Next, the fused features are split, convolved, and passed through a sigmoid activation to generate the CoordAtt coefficients $a_i^h\in \mathbb{R}^{C\times H\times1}$ and $a_i^w\in \mathbb{R}^{C\times 1\times W}$ for the height and width directions, respectively. Finally, the input feature $F_i$ is re-weighted through a Hadamard product with the computed attention coefficients $a_i^h$ and $a_i^w$ to perform the weighting operation.
\begin{align}
    x_i^h &= \text{AvgPool}_w(F_i), \quad x_i^w = \text{AvgPool}_h(F_i),\\
    f_i &= \delta(\text{BN}(\text{Conv}_f(\text{Concat}[x_i^h, (x_i^w)^T]))), \\
    F^{\text{branch2}}_i &= F_i \odot \sigma(\text{Conv}_h(f^h_i)) \odot \sigma(\text{Conv}_w(f^w_i)),
\end{align}
where $\text{AvgPool}_w$ and $\text{AvgPool}_h$ are 1D average pooling operations. $Concat$ denotes concatenation along the spatial dimension, $\text{Conv}_f$, $\text{Conv}_h$, and $\text{Conv}_w$ are separate 1×1 convolutions, $\delta$ is ReLU, and $\sigma$ is Sigmoid. The fused tensors $f_i$ are divided into $f_i^h$ and $f_i^w$ to generate directional attention maps.

The resultant feature maps from both pathways are then integrated through channel-wise concatenation and a subsequent convolutional fusion, yielding the final output representation:
\begin{equation}
    F^{\text{final}}_i =\text{Conv}_{1\times 1}(\text{Concat}[F^{\text{branch1}}_i, F^{\text{branch2}}_i]).  \label{eq:fused}
\end{equation}
\subsection{Lightweight Gated Fusion Unit (LGFU)}
After the previous two steps, the multi-scale difference features are obtained through element-wise subtraction. The subsequent challenge lies in effectively fusing them to generate a final change feature map that incorporates both deep semantic information and shallow spatial details. Deliberately bypassing the intricate designs in feature decoders, we introduce a LGFU that establishes cross-scale information ``flow'' to aggregate both global context and local details. The fundamental principle of the LGFU is to selectively fuse features in the decoder's upsampling path. Instead of naively merging high-level semantic features with all available low-level details—which are often contaminated by noise and irrelevant clutter—the LGFU introduces a learnable gating mechanism. This gate adaptively controls the proportion of shallow-feature information that flows through at a pixel-wise level, effectively filtering out noise while preserving the integrity of meaningful change boundaries.

Specifically, given the $i$-th layer's deep difference feature $F_{\text{deep}} \in \mathbb{R}^{C_i \times (H/2^i) \times (W/2^i)}$, it first undergoes $2\times$ upsampling and a $1 \times 1$ convolution to align with the adjacent difference feature, $F_{\text{shallow}} \in \mathbb{R}^{C_{i-1} \times (H/2^{i-1}) \times (W/2^{i-1})}$, in terms of spatial resolution and channel dimension, respectively. Subsequently, the aligned deep features and the shallow features are concatenated along the channel dimension and then fed into the Gating Unit (GU). The GU consists of a $3 \times 3$ convolution, followed by Batch Normalization, a ReLU activation, a $1 \times 1$ convolution, and a terminal Sigmoid function, which outputs a single-channel gating map, $G_{map} \in \mathbb{R}^{1 \times (H/2^{i-1}) \times (W/2^{i-1})}$, with pixel values normalized to the range $[0, 1]$. The gating map $G_{map}$ then modulates the shallow features $F_{\text{shallow}}$ pixel-wise by performing a Hadamard product, where the single-channel $G_{map}$ is broadcast across all channels of $F_{\text{shallow}}$ via Python's broadcasting mechanism. This operation ``gates'' the model to learn at each spatial location how much fine-grained detail from the shallow layer should be passed through to refine the semantic predictions from the deep layer. Ultimately, the gated shallow features $F_{\text{shallow}}$ are incorporated into the aligned deep features $F'_{\text{deep}}$ using a residual-like addition. This can be expressed as:
\begin{align}
    F'_{\text{deep}} &= \text{Conv}_{1 \times 1}(\text{Upsample}(F_{\text{deep}})), \\
    G_{map} &= \text{GU}(\text{Concat}[F'_{\text{deep}}, F_{\text{shallow}}]), \\
    F'_{\text{fused}} &= G_{map} \odot F_{\text{shallow}} + F'_{\text{deep}}.
\end{align}

This fusion process is iteratively applied in a bottom-up manner across the decoder, progressively integrating features from shallower layers until the final high-resolution change map is generated.

Finally, the entire network is optimized end-to-end using a composite loss function. To address the pervasive class imbalance in CD, we combine the Binary Cross-Entropy ($\mathcal{L}_{\text{BCE}}$) and Dice ($\mathcal{L}_{\text{Dice}}$) losses, leveraging the former's ability to enforce pixel-level accuracy and the latter's robustness to skewed distributions~\cite{2024hybrid,WS-Net++}. The final objective function is a linear combination of the two:
\begin{equation}
\mathcal{L}_{\text{Total}} = \mathcal{L}_{\text{BCE}} + \mathcal{L}_{\text{Dice}}.
\label{eq:total_loss}
\end{equation}
\section{Experiments and Analysis}\label{s4}
\label{sec:experiments}
\subsection{Datasets}
\label{subsec:datasets}
Comparative experiments were conducted on three publicly available datasets:
\begin{enumerate}
     \item {CDD}~\cite{CDD} dataset, captured from Google Earth, contains 11 pairs of raw images with varied resolutions (0.03 to 1 m/pixel), comprising both seasonal and disaster-induced changes. This dataset is particularly challenging due to complex scenarios involving shadows, lighting variations, and cloud cover \cite{wang2025enhancing}. We used the official pre-processed version, which consists of 10,000, 3,000, and 3,000 image patches of size 256$\times$256 for the training, validation, and test sets, respectively.
    \item {GZ-CD}~\cite{GZ-CD} dataset, sourced from Google Earth via BIGEMAP, comprises season-varying, high-resolution (0.55 m/pixel) image pairs of Guangzhou's suburbs. The original images were cropped into 256×256 non-overlapping patches, resulting in 1073 clips (originally 1067 in the source paper) after filtering out those containing changed pixels. The final training, validation, and test sets were split with 751, 214, and 108 pairs, respectively.
    \item {LEVIR-CD}~\cite{LEVIR-CD} dataset, a large-scale benchmark for building CD, consists of 637 image pairs (0.5 m/pixel)  sized at 1024$\times$1024. These images document diverse building changes, such as the emergence of villas and warehouses, over a 5 to 14-year span in multiple cities across Texas, USA. Following the official protocol, the images are partitioned into 7,120, 1,024, and 2,048 patches of 256$\times$256 for training, validation, and testing, respectively.
\end{enumerate}
\subsection{Experimental Setup}
\label{subsec:Setup}
 \subsubsection{Implementation Details} All experiments were conducted in PyTorch on a single NVIDIA RTX 4090 GPU. We optimized the network using the AdamW optimizer with a weight decay of 0.01. The initial learning rate was set to 1e-3 for the randomly initialized decoder and head, and 1e-4 for the pre-trained backbone, both decayed to 1e-6 over 100 epochs via a cosine annealing schedule. A batch size of 32 was used for all datasets.
 \subsubsection{Evaluation Metrics} Quantitative evaluation is conducted using five standard metrics: Precision (Pre.), Recall (Rec.), F1-score (F1), Intersection over Union (IoU), and Overall Accuracy (OA):
\begin{align}
    \text{Pre.} &= \frac{\text{TP}}{\text{TP} + \text{FP}}, \\
    \text{Rec.} &= \frac{\text{TP}}{\text{TP} + \text{FN}}, \\
    \text{F1} &= \frac{2 \times \text{Pre} \times \text{Rec}}{\text{Pre} + \text{Rec}}, \\
    \text{IoU} &= \frac{\text{TP}}{\text{TP} + \text{FP} + \text{FN}}, \\
    \text{OA} &= \frac{\text{TP} + \text{TN}}{\text{TP} + \text{FP} + \text{TN} + \text{FN}},
\end{align}
where TP, FP, TN, and FN represent the counts of true positives, false positives, true negatives, and false negatives. We prioritize the F1 and IoU for providing a more balanced evaluation.

\begin{figure*}[t]
    \centering
    \includegraphics[width=\textwidth]{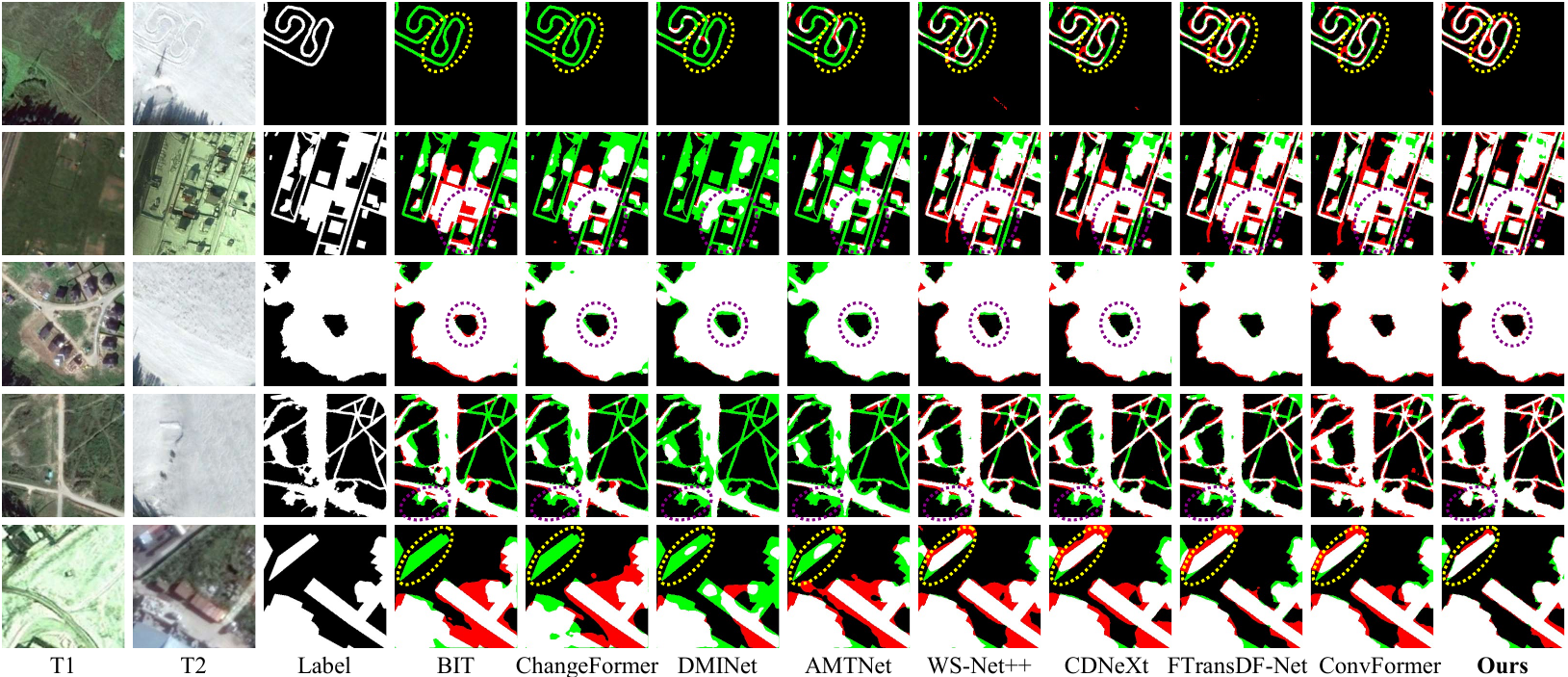}
    \caption{Visualization of experimental results on the CDD dataset. White indicates true positives, \textbf{black} signifies true negatives, \tcr{red} marks false positives, and \tcg{green} highlights false negatives. Colored dashed boxes highlight the differences among the algorithms.}
    \label{CDDVisualization}
\end{figure*}

\subsection{Comparisons With SOTAs}
\label{subsec:SOTAs}
To validate the effectiveness and efficiency of FSG-Net, widely adopted SOTA methods used as benchmarks for bi-temporal CD are implemented and compared, including the following.
\begin{enumerate}
    \item {BIT}~\cite{BIT} pioneers a hybrid CNN-Transformer architecture, where a Transformer encoder leverages SelfAtt on CNN feature to capture long-range dependencies.
    \item {ChangeFormer}~\cite{Changeformer} employs a Siamese hierarchical Transformer encoder and an MLP decoder to capture multi-scale, long-range contextual details.
    \item {DMINet}~\cite{DMINet} constructs an intertemporal joint-attention module for early feature interaction, enabling guided, dual-branch difference acquisition.
    \item {AMTNet}~\cite{ISPRSattention} utilizes an integrated CNN-Transformer for feature exchange-based domain alignment and attention-driven multi-scale feature enhancement.
    \item {WS-Net++}~\cite{WS-Net++} introduces a wavelet Siamese network that uses semi-supervised domain adaptation to mitigate appearance shifts and reduce pseudo-changes.
    \item {CDNeXt}~\cite{CDNeXt} proposes a Temporospatial Interactive Attention to explicitly model and correct for geometric perspective rotation and temporal style differences.
    \item {FTransDF-Net}~\cite{FTransDF-Net} features a dual architecture with a gated module for multi-scale detail fusion and a frequency Transformer for long-range dependency capture.
    \item {ConvFormer}~\cite{ConFormer-CD} integrates parallel convolution, SelfAtt, and a novel Temporal Attention mechanism for difference-guided cross-temporal relationship modeling.
\end{enumerate}    

Above compared models were benchmarked using their publicly available implementations and default parameters.
\subsection{Experimental Results}
\subsubsection{Quantitative Evaluation} Table~\ref{tab:sota_comparison_with_venue} summarizes the quantitative comparison on the CDD, GZ-CD, and LEVIR-CD datasets, where it is evident that the proposed FSG-Net attains superior results on nearly every metric. 
\begin{figure*}[t]
    \centering
    \includegraphics[width=\textwidth]{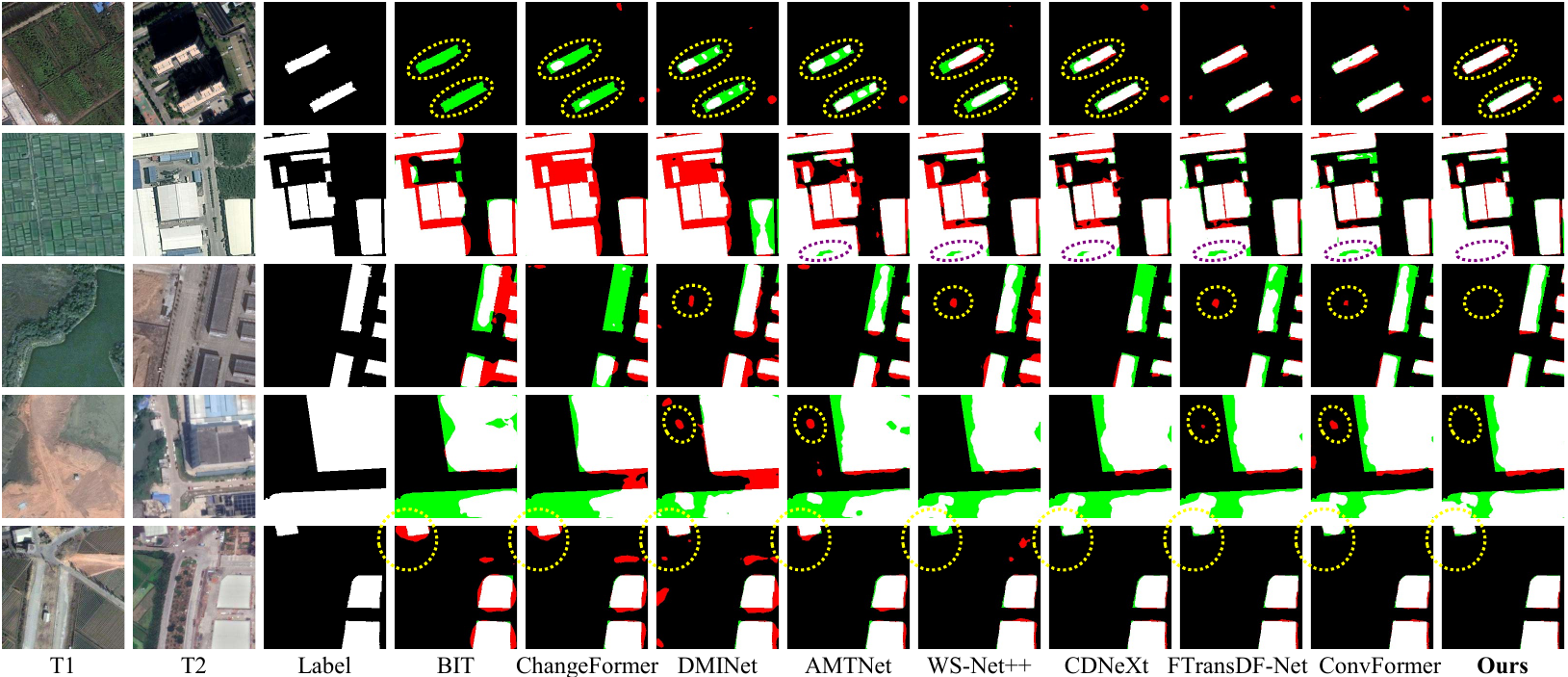}
    \caption{Visualization of experimental results on the GZ-CD dataset. White indicates true positives, \textbf{black} signifies true negatives, \tcr{red} marks false positives, and \tcg{green} highlights false negatives. Colored dashed boxes highlight the differences among the algorithms.}
    \label{GZCDVisualization}
\end{figure*}
\begin{table*}[!t]
\centering
\caption{COMPARISON RESULTS ON THE THREE CD DATASETS. THE BEST SCORE IS HIGHLIGHTED IN \textbf{BOLD}. \\ALL RESULTS ARE DESCRIBED AS PERCENTAGES (\%).}
\label{tab:sota_comparison_with_venue}
\small
\setlength{\tabcolsep}{2.5pt}
\renewcommand{\arraystretch}{1.2}
\begin{tabular}{c|c|ccccc|ccccc|ccccc} 
\toprule
\multirow{2.5}{*}{\textbf{Methods}} & \multirow{2.5}{*}{\textbf{Ref.}} & \multicolumn{5}{c|}{\textbf{CDD}} & \multicolumn{5}{c|}{\textbf{GZ-CD}} & \multicolumn{5}{c}{\textbf{LEVIR-CD}} \\
\cmidrule(lr){3-7} \cmidrule(lr){8-12} \cmidrule(lr){13-17}

& & Pre. & Rec. & F1 & IOU & OA & Pre. & Rec. & F1 & IOU & OA & Pre. & Rec. & F1 & IOU & OA \\ 
\midrule
BIT~\cite{BIT}            & TGRS21 & 90.67 & 86.44 & 88.50 & 79.37 & 97.26 & 82.90 & 80.68 & 81.77 & 69.16 & 90.63 & 89.24 & 89.37 & 89.30 & 80.67 & 98.88 \\
ChangeFormer~\cite{Changeformer} & IGARSS22 & 91.54 & 89.31 & 90.41 & 82.50 & 97.68 & 84.59 & 84.28 & 84.43 & 73.06 & 91.90 & 91.97 & 89.03 & 90.48 & 82.62 & 99.02 \\
DMINet~\cite{DMINet}      & TGRS23 & 92.45 & 90.72 & 91.58 & 84.47 & 97.96 & 89.04 & 86.52 & 87.76 & 78.19 & 93.71 & 92.27 & 88.94 & 90.57 & 82.77 & 99.03 \\
AMTNet~\cite{ISPRSattention}          & ISPRS23 & 92.32 & 92.91 & 92.61 & 86.24 & 98.19 & 88.51 & 85.27 & 86.86 & 76.77 & 93.28 & 91.34 & 89.76 & 90.54 & 82.72 & 99.02 \\
WS-Net++~\cite{WS-Net++}      & TGRS24 & 92.95 & 93.21 & 93.08 & 87.06 & 98.31 & 88.32 & 85.76 & 87.02 & 77.02 & 93.33 & 92.11 & \textbf{90.05} & 91.07 & 83.60 & 99.08 \\
CDNeXt~\cite{CDNeXt}      & JAG24 & 92.76 & 93.87& 93.31 & 87.46 &98.35  & 89.21 & 86.34 & 87.75 & 78.17 & 93.72 & 92.15 & 89.69 & 90.91 & 83.33 & 99.06 \\
FTransDF-Net~\cite{FTransDF-Net}    & JAG25 & 93.10 & 94.03 & 93.56 & 87.90 & 98.42 &89.13 &87.45  & 88.28 & 79.02 & 93.95& 92.19 &89.73  & 90.94 & 83.39&99.07  \\
ConvFormer~\cite{ConFormer-CD}   & TGRS25 & \textbf{93.35} & 93.19 & 93.28 & 87.41 & 98.36 & 89.11 & 87.58 &88.33 & 79.11 & 93.97 &92.14  & 89.69 & 90.90 & 83.32 & 99.06\\
\midrule
\textbf{FSG-Net} & Ours & 93.33 & \textbf{95.01} & \textbf{94.16} & \textbf{88.96} & \textbf{98.56} & \textbf{90.85} & \textbf{88.20} & \textbf{89.51} & \textbf{81.01} & \textbf{94.61} & \textbf{92.53} & 90.04 & \textbf{91.27} & \textbf{83.94} & \textbf{99.10} \\
\bottomrule
\end{tabular}
\end{table*}
\begin{figure*}[t]
    \centering
    \includegraphics[width=\textwidth]{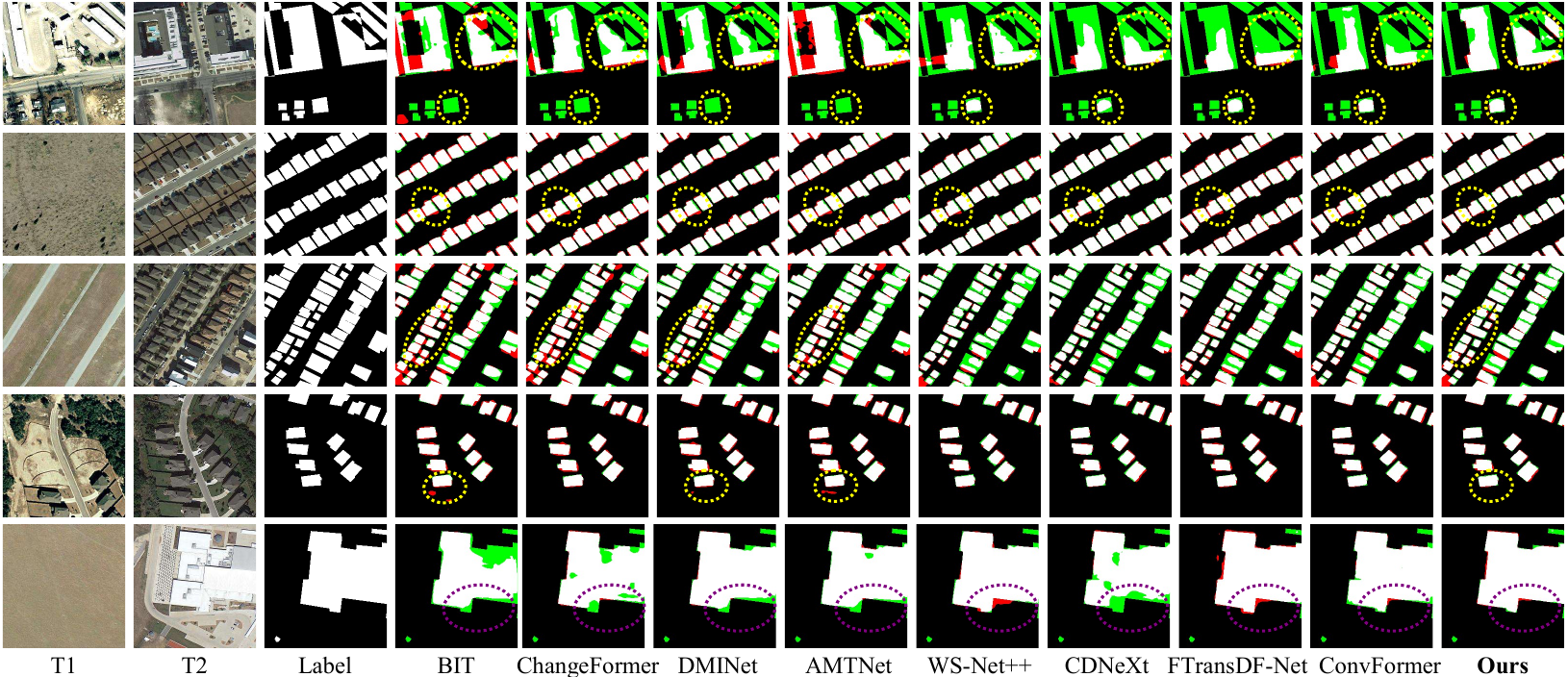}
    \caption{Visualization of experimental results on the LEVIR-CD dataset. White indicates true positives, \textbf{black} signifies true negatives, \tcr{red} marks false positives, and \tcg{green} highlights false negatives. Colored dashed boxes highlight the differences among the algorithms.}
    \label{levirVisualization}
\end{figure*}
Overall, FSG-Net achieves an F1 score of approximately 90\% across all three datasets. A noticeable performance degradation for all methods is observed on the GZ-CD dataset. This is likely a consequence of two compounding factors: first, the data scarcity hinders the models from learning the complex change patterns robustly. Second, the dataset's high prevalence of seasonal discrepancies introduces significant ambiguity, further complicating the detection task with limited training data. Even under these demanding conditions, the proposed algorithm still exhibits strong performance, outperforming all comparison models with a Pre. of 90.85\%, Rec. of 88.20\%, F1 of 89.51\%, IoU of 81.01\%, and OA of 94.61\%. This remarkable performance stems from the synergistic interplay between the DAWIM and STSAM. The former mitigates spurious variations by operating in the frequency domain, while the latter amplifies attention towards salient change regions. This collaborative mechanism results in a marked increase in the model's sensitivity to true changes. Likewise, the competitive performance of FSG-Net extends to the CDD and LEVIR-CD datasets, where it surpasses the recent ConvFormer by 1.55\% and 0.62\% in IoU, respectively.

\subsubsection{Qualitative Evaluation} Qualitatively, Figs. \ref{CDDVisualization} to \ref{levirVisualization} present the predicted maps across different datasets, with distinct colors used to indicate the detection results: TP (white), TN (black), FP (red), and FN (green), highlighting the correctness or errors in the predictions. Colored dashed boxes highlight the differences among the algorithms.

Visualization on CDD (Fig.~\ref{CDDVisualization}): As shown in the 1st and 4th rows, most algorithms struggle with the complex, crisscrossing network of rural paths, leading to either missed detections (e.g., BIT) or false alarms (e.g., WS-Net++). Impressively, FSG-Net demonstrates exceptional resilience in this complex scene. It almost flawlessly traces the full extent of the path's change with high fidelity, notwithstanding a few negligible and inevitable misclassifications. During the construction of new agricultural facilities on barren land (5th row), severe illumination artifacts induce most algorithms to generate large, contiguous regions of pseudo-changes. This vulnerability is precisely what FSG-Net addresses through its differential frequency-domain interaction, which is designed to disentangle semantic changes from such non-semantic, illumination-induced noise. Collectively, the aforementioned examples provide compelling evidence of FSG-Net proficiency in both suppressing pseudo-changes and capturing genuine change targets.

Visualization on GZ-CD (Fig.~\ref{GZCDVisualization}): A closer look at the 1st row reveals that in the scenario where two new buildings are constructed on farmland, FSG-Net not only segments the entire building framework completely but also preserves the boundary integrity with remarkable clarity at the transition between the foreground (changed) and background (unchanged) regions. Such superior performance is intrinsically linked to the contribution of the STSAM. On one hand, the CrossAtt, augmented with temporal embeddings, models long-range contextual relationships across time between the bi-temporal features. On the other hand, the CoordAtt further enhances the focus on the fine-grained details of target edges. In this manner, changes, irrespective of their scale (global or local), are given due consideration and are not overlooked.

Visualization on LEVIR-CD (Fig.~\ref{levirVisualization}): The dataset, specifically labeled for building CD, is characterized by a wide spectrum of change scenes. It features changes ranging from the emergence of large-scale industrial structures, such as factories and warehouses, to the construction of dense residential clusters. The morphological heterogeneity of these structures presents a substantial hurdle for accurate boundary delineation. Owing to the LGFU, which narrows the semantic gap between high-level and low-level features, FSG-Net maintains the integrity of building structures and the sharpness of their edges more effectively than all competing algorithms. Consistent with the quantitative metrics in Table~\ref{tab:sota_comparison_with_venue}, the qualitative analysis further underscores the SOTA capabilities of FSG-Net on the LEVIR-CD.

\subsection{Ablation Experiment and Analysis}
To provide a structured and meaningful evaluation, we benchmark against a competitive baseline, incorporating proven technologies like basic frequency differencing and SelfAtt (replacing DAWIM and STSAM respectively)~\cite{frequency1,self-attention,BIT} with a U-Net decoder. This strong baseline serves to demonstrate that the performance gains of FSG-Net are not merely due to the inclusion of generic enhancements, but stem from specific, refined designs. Table~\ref{tab:ablation_study} presents the ablation results for each component, focusing on the F1 and IoU.
\begin{table}[!t]
\centering
\caption{ABLATION STUDY OF DIFFERENT MODULES. ALL RESULTS ARE DESCRIBED AS PERCENTAGES (\%).}
\label{tab:ablation_study}
\renewcommand{\arraystretch}{1.2} 
\setlength{\tabcolsep}{3pt}      

\begin{tabular}{ccc cc cc cc} 
\toprule

\multirow{2.5}{*}{DAWIM} & \multirow{2.5}{*}{STSAM} & \multirow{2.5}{*}{LGFU} & \multicolumn{2}{c}{CDD} & \multicolumn{2}{c}{GZ-CD} & \multicolumn{2}{c}{LEVIR-CD} \\
\cmidrule(lr){4-5} \cmidrule(lr){6-7} \cmidrule(lr){8-9} 
& & & F1 & IOU & F1 & IOU & F1 & IOU \\
\midrule
\multicolumn{3}{c}{Baseline} & 90.33 & 82.37 & 84.97 & 73.87 & 90.29 & 82.30 \\
\midrule 
$\checkmark$ & $\times$ & $\times$ & 91.97 & 85.13 & 86.80 & 76.68 & 90.45 & 82.57 \\
$\times$ & $\checkmark$ & $\times$ & 91.60 & 84.50 & 86.55 & 76.29 & 90.71 & 83.00 \\
$\times$ & $\times$ & $\checkmark$ & 91.15 & 83.74 & 85.80 & 75.13 & 90.62 & 82.85 \\
$\times$ & $\checkmark$ & $\checkmark$ & 92.50 & 86.05 & 87.73 & 78.14 & 91.05 & 83.57 \\
$\checkmark$ & $\times$ & $\checkmark$ & 92.88 & 86.71 & 87.95 & 78.49 & 90.73 & 83.03 \\
$\checkmark$ & $\checkmark$ & $\times$ & 93.35 & 87.53 & 88.57 & 79.48 & 91.01 & 83.50 \\
\midrule
\multicolumn{3}{c}{\textbf{Ours}} & \textbf{94.16} & \textbf{88.96} & \textbf{89.51} & \textbf{81.01} & \textbf{91.27} & \textbf{83.94} \\
\bottomrule
\end{tabular}
\end{table}
\begin{figure}[t!]
    \centering
    \includegraphics[width=1.0\columnwidth]{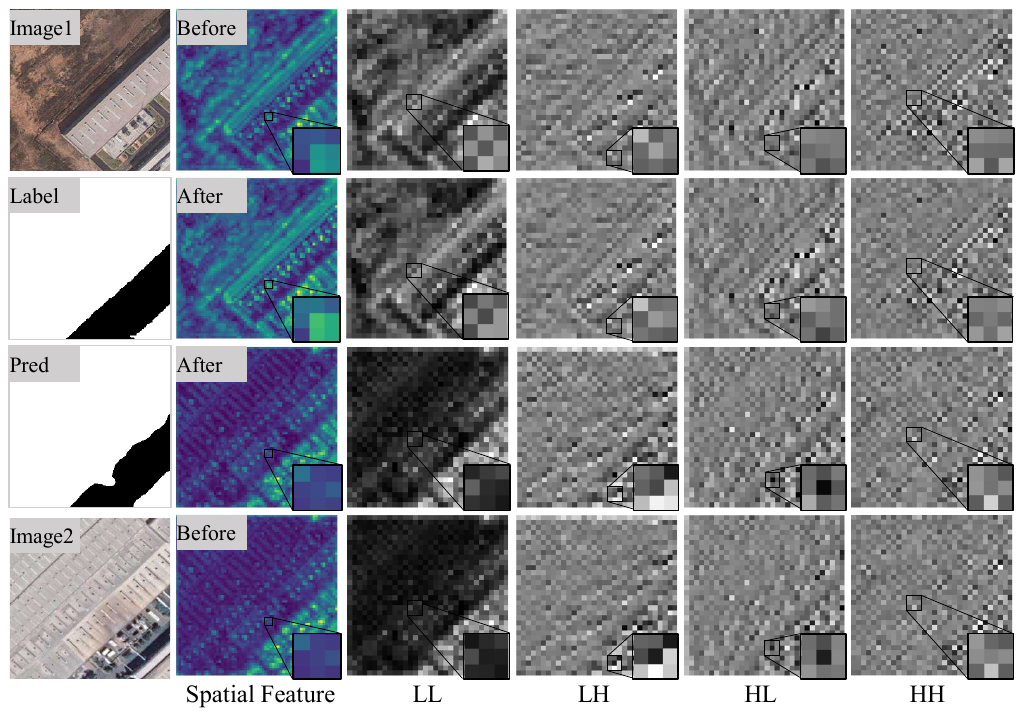}
    \caption{Visualization of the DAWIM's feature modulation process. \textbf{Rows 1 \& 4} show the features before interaction for Image1 and Image2. \textbf{Rows 2 \& 3} show the features after interaction. The ground truth change mask (Label) and the model's final prediction (Pred) are also provided for reference. The columns display the input image, the spatial feature map, and its wavelet components.}
    \label{dawim}
\end{figure}
\begin{figure}[t!]
    \centering
    \includegraphics[width=1.0\columnwidth]{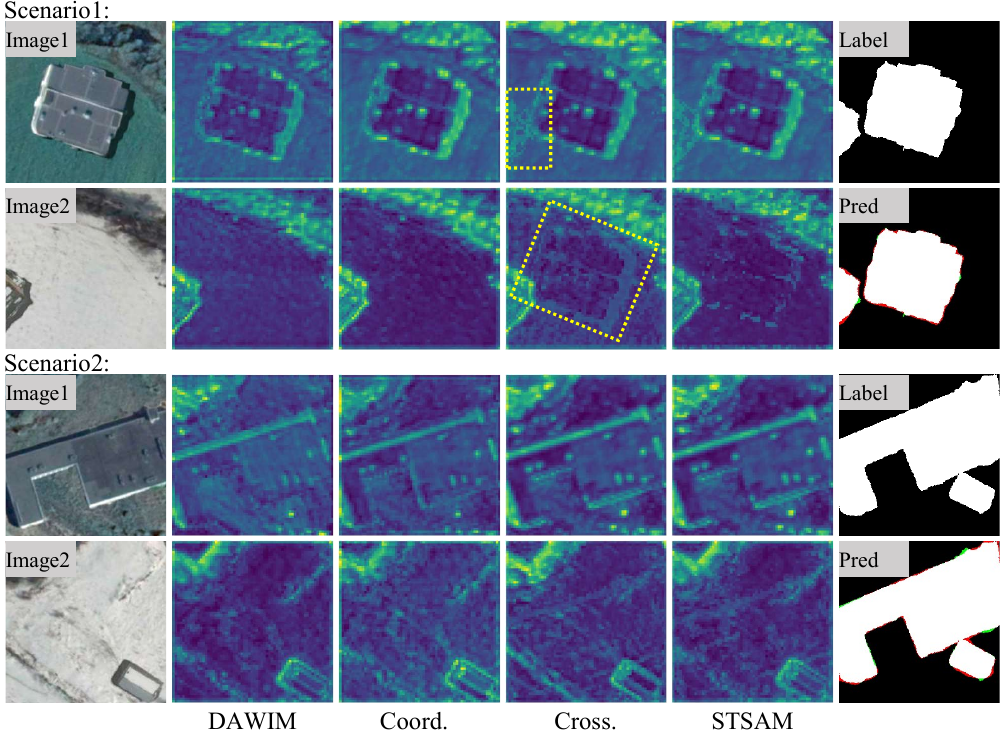}
    \caption{Visualization of the progressive feature refinement process within the STSAM across two challenging scenarios. \textbf{Columns} from left to right show: the input images (Image1, Image2), and the feature heatmaps after the DAWIM, Coordinate Attention (Coord.), Cross-Attention (Cross.), and the final STSAM stage. For each scenario, the ground truth (Label) and the final prediction (Pred) are provided for comparison. The yellow boxes highlight the reciprocal attention focusing mechanism of the Cross-Attention.}
    \label{stsam}
\end{figure}
\subsubsection{Effectiveness of DAWIM} The primary role of the DAWIM is to suppress pseudo-changes by discerningly processing features in the frequency domain. To validate this, we isolated its effect by adding it to the Baseline. As shown in Table~\ref{tab:ablation_study} (row 2 vs. row 1), this integration yields a consistent performance boost across all datasets. Most notably, it achieves F1 improvements of 1.64\% on CDD and 1.83\% on GZ-CD. The fact that these substantial gains occur on datasets characterized by significant seasonal variations strongly corroborates the exceptional capability of the DAWIM in mitigating task-irrelevant interference. This confirms that the strategy of applying distinct, tailored interactions to different frequency components is more effective than a simple, uniform differencing approach. Observing the results in rows 1, 2, 3, and 7, it is evident that combining DAWIM and STSAM yields a significant performance gain over their individual contributions. This reciprocal effect can be attributed to a sequential enhancement mechanism: DAWIM's initial filtering of background noise allows the STSAM to subsequently focus its temporal-spatial attention mechanism on authentic change targets with greater precision, thereby achieving a cumulative benefit greater than the sum of its parts.

To provide a more intuitive understanding of the DAWIM, as shown in Fig.~\ref{dawim}, we visualize the normalized internal feature representations before and after the frequency-domain interaction. The figure is structured in a ``sandwich'' layout, where the features maps before interaction (Rows 1 and 4 for T1 and T2 respectively) flank after interaction (Rows 2 and 3). From the magnified insets highlighting feature differences before and after interaction, we can draw two key observations. First, the module effectively preserves crucial spatial structures, as evidenced by the overall consistency in feature patterns before and after the interaction. Second, it successfully introduces temporal dynamics, indicated by the subtle yet distinct changes in feature intensity and texture post-interaction, an effect that is quantitatively supported by the results in Table~\ref{tab:ablation_study}. 
\begin{figure}[t!]
    \centering
    \includegraphics[width=1.0\columnwidth]{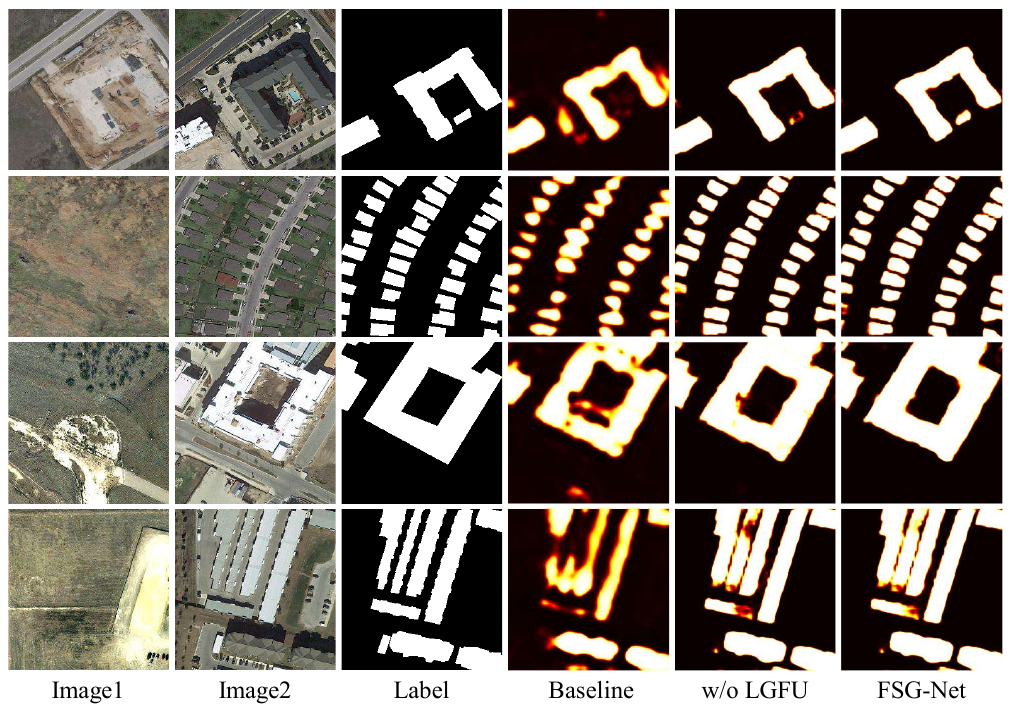}
    \caption{Visualization of the LGFU's contribution to refining prediction confidence heatmaps. \textbf{Columns} from left to right show: the bi-temporal images (Image1, Image2), the ground truth (Label), and the predicted heatmaps from Baseline, the model without LGFU (w/o LGFU), and the complete FSG-Net. The color intensity corresponds to the predicted change probability, ranging from black (low) to white (high).}
    \label{lgfu}
\end{figure}

\subsubsection{Effectiveness of STSAM} The STSAM is designed to refine the representation of genuine changes by creating an information flow pathway between bi-temporal features to mutually steer the attention distribution effectively. As detailed in the rows 6 and 8 of Table~\ref{tab:ablation_study}, ablating the STSAM module leads to a drop in F1 of 1.28\%, 1.56\%, and 0.54\% on the CDD, GZ-CD, and LEVIR-CD datasets, respectively, compared to the full model. By capturing temporal dependencies globally via CrossAtt augmented with temporal embedding and encoding precise spatial locations locally through CoordAtt, the STSAM ensures that genuinely changed regions, irrespective of their scale or complexity, are saliently represented, thereby achieving complementary temporal-spatial focus on change targets.

Fig.~\ref{stsam} illustrates the progressive feature refinement by the STSAM in two representative scenarios featuring building changes amidst seasonal shifts. Evidently, the CoordAtt branch sharpens the feature representation, accentuating fine-grained spatial details and preserving the integrity of building boundaries. The yellow-boxed regions in the fourth column warrant special attention as they highlight a key observation: through the temporally-enhanced CrossAtt, the T1 features perceive the new building construction in the T2 image, prompting a heightened focus on the counterpart location within itself. Conversely, the T2 features identify the demolition in T1 and similarly amplify attention on their own counterpart region. The final fused feature map showcases the synergy of these two streams: it inherits the high-level semantic focus from the temporal branch while being refined with the high-fidelity structural information from the spatial branch. This fusion enables the model to robustly identify multi-scale changes while preserving sharp boundary integrity, as evidenced in the final segmentation results. The bottom two rows present a similar yet more challenging case where a new building is camouflaged by snow cover, making it visually blend with the environment. Remarkably, due to the robust frequency-spatial interplay, FSG-Net precisely delineates the changed area despite these challenging camouflage-like artifacts, a result clearly visible in the final prediction.
\subsubsection{Effectiveness of LGFU} Bridging the semantic gap between deep abstract features and shallow detail-rich features is another critical challenge for accurate CD. The LGFU achieves this by leveraging high-level semantic information to guide and enrich shallow features. A clear observation from Table~\ref{tab:ablation_study} reveals the significant contribution of the LGFU. Integrating the LGFU individually into the baseline boosts the IoU by 1.37\%, 1.26\%, and 0.55\% on the respective datasets. More revealingly, when LGFU is the sole module ablated from the full model, the performance degradation is even more pronounced, with IoU dropping by 1.43\% and 1.53\% on the first two datasets. This asymmetry is a crucial finding: It indicates that the LGFU's contribution is amplified when it operates on the highly refined features generated by DAWIM and STSAM. This validates a powerful synergistic effect within our architecture, where the modules not only contribute individually but also enhance each other's efficacy, culminating in a final model that is greater than the sum of its parts.

\begin{table}[!t]
\centering
\caption{ANALYSIS ON DAWIM'S INTERACTION STRATEGIES. \\THE METRIC IS F1 SCORE (\%).}
\label{tab:ablation_dawim}
\renewcommand{\arraystretch}{1.2}
\setlength{\tabcolsep}{1pt} 

\newcolumntype{C}[1]{>{\centering\arraybackslash}p{#1}}

\newcolumntype{L}[1]{>{\raggedright\arraybackslash}p{#1}}

\begin{tabular}{C{1.4cm} C{1.2cm} C{1.2cm} C{1.35cm} C{1.35cm} C{1.4cm} } 
\toprule
Strategy & Weighted & Residual & CDD & GZ-CD & LEVIR-CD \\ 
\midrule
Difference & $\checkmark$ & $\checkmark$ & 92.83 & 88.03 & 91.09 \\ 
Conv$_{2 \times 1 \times 1}$ & $\checkmark$ & $\checkmark$ & 93.47 & 88.42 & 91.12 \\ 
Conv$_{2 \times 3 \times 3}$ & $\checkmark$ & $\checkmark$ & 93.65 & 88.92 & 91.17 \\ 
w/o SE & $\times$ & $\checkmark$ & 93.97 & 89.28 & 91.20 \\ 
w/o Res & $\checkmark$ & $\times$ & 92.57 & 87.81 & 90.94 \\ 
\midrule
DAWIM & $\checkmark$ & $\checkmark$ & \textbf{94.16} & \textbf{89.51} & \textbf{91.27} \\ 
\bottomrule
\end{tabular}
\end{table}

Fig.~\ref{lgfu} visualizes the LGFU's contribution using a predicted probability heatmap, where the color of each pixel maps to the model's confidence score for the change class. The color intensity signifies the probability of change, ascending from black (lowest) through red and yellow to white (highest). It is easily observed that the baseline captures the general change regions, yet it struggles with blurry boundaries and noisy interferences. The predominantly yellow prediction map further suggests a lack of absolute certainty in its identification of changed areas. Only upon the final integration of the LGFU does the model produce exceptionally sharp and well-defined change boundaries that closely adhere to the ground truth. This provides compelling visual evidence that the LGFU is instrumental not just in bridging the semantic gap, but ultimately in translating high-level conceptual understanding into pixel-perfect boundary delineations.
\subsubsection{Analysis on DAWIM's Interaction Strategies} To validate the internal design of DAWIM, we compared different frequency-domain interaction strategies, including: unified approaches (e.g., Difference, Conv2x1x1, Conv2x3x3) for all components, and the removal of the SE weighting mechanism and the residual connection, as detailed in Table~\ref{tab:ablation_dawim}. Taking the CDD as an example, the DAWIM surpasses the other three unified strategies in F1 by 1.33\%, 0.69\%, and 0.51\% respectively, indicating that a tailored interaction strategy could maximize the benefits of frequency-domain feature interaction. Another noteworthy observation is that removing the residual connection leads to a catastrophic collapse, with the F1 dropping by 1.59\%, 1.70\%, and 0.33\% on the three datasets, respectively. It can be attributed to a well-known consensus in deep learning: removing residual connections hinders model convergence and impedes the smooth flow of gradients and information~\cite{Attention}. This suggests that a robust and easily optimizable architecture is a prerequisite for fully unleashing the potential of advanced frequency-domain interaction strategies.
\subsubsection{Comparison on STSAM's Components}
Table~\ref{tab:ablation_stsam} presents an in-depth ablation study, comparing different standard attention mechanisms and the effect of removing the temporal embedding from the complete STSAM. 
\begin{table}[!t]
\centering
\caption{COMPARISON ON STSAM'S DIFFERENT COMPONENTS.\\THE METRIC IS F1 SCORE (\%).}
\label{tab:ablation_stsam}
\renewcommand{\arraystretch}{1.2}
\setlength{\tabcolsep}{5pt}

\newcolumntype{C}[1]{>{\centering\arraybackslash}p{#1}}

\begin{tabular}{c C{1.4cm} C{1.4cm} C{1.4cm}} 
\toprule
Component & CDD & GZ-CD & LEVIR-CD \\ 
\midrule
Cross-Attention & 93.69 & 88.93 & 91.02 \\ 
Coord-Attention & 92.36 & 87.41 & 90.31 \\ 
Self-Attention  & 92.88 & 87.95 & 90.73 \\ 
Self \& Coord-Attention & 93.47 & 88.74 & 90.91 \\ 
w/o time embedding    & 94.01 & 89.34 & 91.18 \\ 
\midrule
STSAM & \textbf{94.16} & \textbf{89.51} & \textbf{91.27} \\ 
\bottomrule
\end{tabular}
\end{table}
Substituting SelfAtt with CoordAtt slightly decreases detection performance, due to the latter's inability to effectively capture long-range dependencies. When the two attention mechanisms are combined, the global modeling capabilities and direction-aware feature encoding yield a certain degree of performance improvement. Nevertheless, the inherent functional overlap in this combination keeps the F1 lower by 0.22\%, 0.19\%, and 0.11\% on the three datasets, respectively, compared to using only CrossAtt. The last two rows reveal that explicitly amplifying temporal disparities between bi-temporal features via time embedding can inject new life into the model, particularly when it hits a performance ceiling, with the F1 increasing by 0.15\%, 0.17\%, and 0.09\%, respectively.

\subsection{Discussion} Synthesizing the quantitative results (Table~\ref{tab:sota_comparison_with_venue}) and qualitative evidence (Figs. \ref{CDDVisualization} to \ref{levirVisualization}), it is evident that FSG-Net effectively suppresses pseudo-changes and bridges the semantic gap through the orchestrated interplay of its DAWIM, STSAM, and LGFU modules. The superiority of our proposed FSG-Net is demonstrated by its SOTA performance across all three public datasets, with an F1 and IoU of 94.16\% and 88.96\% on CDD, 89.51\% and 81.01\% on GZ-CD, and 91.27\% and 83.94\% on LEVIR-CD respectively.  Providing intuitive evidence for this approach, the visualization results in Fig.~\ref{dawim} and Fig.~\ref{stsam} illustrate that DAWIM's frequency-domain interaction suppresses pseudo-changes, while STSAM's spatial interaction enhances genuine changes. This bidirectional interplay mitigates visual feature interference arising from disparate imaging conditions and promotes information coupling within intra-level representations. Moreover, the evidence in Fig.~\ref{lgfu} substantiates the role of the LGFU. It leverages high-level semantics to guide the selection of valuable details from shallow features, resulting in the effective filtering of noise and the generation of clean, well-defined boundaries. However, the current architecture's robustness in extreme scenarios could be further enhanced. A notable limitation, illustrated in the final row of Fig.~\ref{levirVisualization}, is the failure to detect minute, isolated building changes amidst vast homogeneous backgrounds.

Table~\ref{tab:arch_efficiency} compares the architectural complexity and efficiency. The model parameters (Params) and Floating-Point Operations (FLOPs) were based on a unified image input size of 256×256×3.
\begin{table}[!t]
\centering
\caption{COMPARISONS OF MODEL ARCHITECTURE AND EFFICIENCY}
\label{tab:arch_efficiency}
\renewcommand{\arraystretch}{1.2}
\setlength{\tabcolsep}{4.5pt}

\begin{tabular}{cc c c c} 
\toprule
Method & Backbone & \makecell{Frequency \\ Interaction} & Params(M) & FLOPs(G) \\ 
\midrule
BIT & ResNet18 & $\times$ & 3.55 & 4.35 \\
ChangeFormer & MiT-b1 & $\times$ & 13.94 & 52.84 \\
DMINet & ResNet18 & $\times$ & 6.24 & 14.42 \\
AMTNet & ResNet18 & $\times$ & 16.44  & 14.71 \\
WS-Net++ & ResNet18 & $\checkmark$ & -- & -- \\
CDNeXt & CNX-t & $\times$ & 39.42 & 15.76 \\
FTransDF-Net & DFG-UNet & $\checkmark$ & 17.91 & 9.37 \\

\multirow{2}{*}{ConvFormer} & w/ ST \& BT- & \multirow{2}{*}{$\times$} & \multirow{2}{*}{37.72} & \multirow{2}{*}{5.14} \\
& Blocks & & & \\
\midrule
Ours & ResNet18 & $\checkmark$ & 13.76 & 6.21 \\ 
\bottomrule
\end{tabular}
\end{table}
Beyond incorporating an advanced frequency interaction mechanism--a feature absent in most SOTA methods--FSG-Net achieves a superior balance in model complexity. While recent architectures like CDNeXt are parameter-heavy (39.42M) and ChangeFormer is computationally expensive (52.84G), FSG-Net achieves an exceptional trade-off with only 13.76M parameters and 6.21 GFLOPs.

\section{Conclusion}\label{s5}
In this article, we have proposed FSG-Net to address two critical concerns in CD tasks: the high false alarm rates caused by pseudo-changes from varying imaging conditions, and the semantic gap between deep, abstract features and shallow, detail-rich features. For the former, FSG-Net introduces a synergistic frequency-spatial separation strategy. More specifically, it leverages the DAWIM to optimize feature representations by applying tailored processing to different frequency-domain components, thereby suppressing interference from task-irrelevant factors. Subsequently, the STSAM module establishes an information pathway between the bi-temporal features, which in turn mutually guides their attention towards genuine change targets. This synergistic ``push-pull'' mechanism is designed to maximize the benefits of feature interaction. To tackle the second challenge, the LGFU leverages high-level semantics to guide and filter shallow-level details, ensuring that change targets are well-represented regardless of their scale. Extensive experiments on three public datasets demonstrate that FSG-Net delivers SOTA performance, exhibiting a superior trade-off between accuracy and efficiency. 

For future research, our efforts will be directed toward lightweight architectures and integrating multi-modal data or constructing knowledge graphs for rule-based constraints to further enhance the model's capabilities in extreme and complex scenarios.

\bibliographystyle{IEEEtran}
\bibliography{references}

% Generated by IEEEtran.bst, version: 1.14 (2015/08/26)
\begin{thebibliography}{10}
\providecommand{\url}[1]{#1}
\csname url@samestyle\endcsname
\providecommand{\newblock}{\relax}
\providecommand{\bibinfo}[2]{#2}
\providecommand{\BIBentrySTDinterwordspacing}{\spaceskip=0pt\relax}
\providecommand{\BIBentryALTinterwordstretchfactor}{4}
\providecommand{\BIBentryALTinterwordspacing}{\spaceskip=\fontdimen2\font plus
\BIBentryALTinterwordstretchfactor\fontdimen3\font minus \fontdimen4\font\relax}
\providecommand{\BIBforeignlanguage}[2]{{%
\expandafter\ifx\csname l@#1\endcsname\relax
\typeout{** WARNING: IEEEtran.bst: No hyphenation pattern has been}%
\typeout{** loaded for the language `#1'. Using the pattern for}%
\typeout{** the default language instead.}%
\else
\language=\csname l@#1\endcsname
\fi
#2}}
\providecommand{\BIBdecl}{\relax}
\BIBdecl

\bibitem{review1}
D.~Peng, X.~Liu, Y.~Zhang, H.~Guan, Y.~Li, and L.~Bruzzone, ``{Deep learning change detection techniques for optical remote sensing imagery: Status, perspectives and challenges},'' \emph{{Int. J. Appl. Earth Observ. Geoinf.}}, vol. 136, p. 104282, 2025.

\bibitem{wang2024hyperspectral}
M.~Wang, L.~Gao, L.~Ren, X.~Sun, and J.~Chanussot, ``Hyperspectral simultaneous anomaly detection and denoising: Insights from integrative perspective,'' \emph{{IEEE} J. Sel. Top. Appl. Earth Observ. Remote Sens.}, 2024.

\bibitem{wen2019accurate}
X.~Wen, H.~Xie, H.~Liu, and L.~Yan, ``{Accurate reconstruction of the {LoD3} building model by integrating multi-source point clouds and oblique remote sensing imagery},'' \emph{{ISPRS Int. J. Geo-Inf.}}, vol.~8, no.~3, p. 135, 2019.

\bibitem{pang2025special}
L.~Pang, J.~Yao, K.~Li, and X.~Cao, ``{SPECIAL: Zero-shot Hyperspectral Image Classification With {CLIP}},'' \emph{arXiv preprint arXiv:2501.16222}, 2025.

\bibitem{disaster}
N.~Casagli, E.~Intrieri, V.~Tofani, G.~Gigli, and F.~Raspini, ``{Landslide detection, monitoring and prediction with remote-sensing techniques},'' \emph{{Nat. Rev. Earth Environ.}}, vol.~4, no.~1, pp. 51--64, 2023.

\bibitem{cva}
F.~Bovolo and L.~Bruzzone, ``{A theoretical framework for unsupervised change detection based on change vector analysis in the polar domain},'' \emph{{IEEE} Trans. Geosci. Remote Sens.}, vol.~45, no.~1, pp. 218--236, 2006.

\bibitem{rf}
T.~Bai, K.~Sun, S.~Deng, D.~Li, W.~Li, and Y.~Chen, ``{Multi-scale hierarchical sampling change detection using Random Forest for high-resolution satellite imagery},'' \emph{{Int. J. Remote Sens.}}, vol.~39, no.~21, pp. 7523--7546, 2018.

\bibitem{worldview}
J.~Lin, G.~Wang, D.~Peng, and H.~Guan, ``{Edge-guided multi-scale foreground attention network for change detection in high resolution remote sensing images},'' \emph{{Int. J. Appl. Earth Observ. Geoinf.}}, vol. 133, p. 104070, 2024.

\bibitem{2024review}
Q.~Zhu, X.~Guo, Z.~Li, and D.~Li, ``{A review of multi-class change detection for satellite remote sensing imagery},'' \emph{{Geo-Spat. Inf. Sci.}}, vol.~27, no.~1, pp. 1--15, 2024.

\bibitem{31x31}
X.~Ding, X.~Zhang, J.~Han, and G.~Ding, ``{Scaling up your kernels to 31x31: Revisiting large kernel design in cnns},'' in \emph{in Proc. {IEEE/CVF} Conf. Comput. Vis. Pattern Recognit. ({CVPR})}, 2022, pp. 11\,963--11\,975.

\bibitem{Dilated}
F.~Yu and V.~Koltun, ``{Multi-scale context aggregation by dilated convolutions},'' \emph{arXiv preprint arXiv:1511.07122}, 2015.

\bibitem{Poly}
X.~Cai, Q.~Lai, Y.~Wang, W.~Wang, Z.~Sun, and Y.~Yao, ``{Poly kernel inception network for remote sensing detection},'' in \emph{in Proc. {IEEE/CVF} Conf. Comput. Vis. Pattern Recognit. ({CVPR})}, 2024, pp. 27\,706--27\,716.

\bibitem{self-attention}
M.~Hu, C.~Wu, and L.~Zhang, ``{GlobalMind: Global multi-head interactive self-attention network for hyperspectral change detection},'' \emph{{ISPRS J. Photogramm. Remote Sens.}}, vol. 211, pp. 465--483, 2024.

\bibitem{Asymmetric}
X.~Zhang, S.~Cheng, L.~Wang, and H.~Li, ``{Asymmetric cross-attention hierarchical network based on {CNN} and transformer for bitemporal remote sensing images change detection},'' \emph{{IEEE} Trans. Geosci. Remote Sens.}, vol.~61, pp. 1--15, 2023.

\bibitem{Coordinateattention}
Q.~Hou, D.~Zhou, and J.~Feng, ``{Coordinate attention for efficient mobile network design},'' in \emph{in Proc. {IEEE/CVF} Conf. Comput. Vis. Pattern Recognit. ({CVPR})}, 2021, pp. 13\,713--13\,722.

\bibitem{ISPRSattention}
W.~Liu, Y.~Lin, W.~Liu, Y.~Yu, and J.~Li, ``{An attention-based multiscale transformer network for remote sensing image change detection},'' \emph{{ISPRS J. Photogramm. Remote Sens.}}, vol. 202, pp. 599--609, 2023.

\bibitem{DMINet}
Y.~Feng, J.~Jiang, H.~Xu, and J.~Zheng, ``{Change detection on remote sensing images using dual-branch multilevel intertemporal network},'' \emph{{IEEE} Trans. Geosci. Remote Sens.}, vol.~61, pp. 1--15, 2023.

\bibitem{Featureinteraction}
S.~Fang, K.~Li, and Z.~Li, ``{Changer: Feature interaction is what you need for change detection},'' \emph{{IEEE} Trans. Geosci. Remote Sens.}, vol.~61, pp. 1--11, 2023.

\bibitem{sun2025mask}
D.~Sun, J.~Yao, W.~Xue, C.~Zhou, P.~Ghamisi, and X.~Cao, ``{Mask Approximation Net: A Novel Diffusion Model Approach for Remote Sensing Change Captioning},'' \emph{{IEEE} Trans. Geosci. Remote Sens.}, 2025.

\bibitem{FTransDF-Net}
Z.~Li, Z.~Zhang, M.~Li, L.~Zhang, X.~Peng, R.~He, and L.~Shi, ``{Dual Fine-Grained network with frequency Transformer for change detection on remote sensing images},'' \emph{{Int. J. Appl. Earth Observ. Geoinf.}}, vol. 136, p. 104393, 2025.

\bibitem{frequency1}
Y.~Tang, S.~Feng, C.~Zhao, Y.~Fan, Q.~Shi, W.~Li, and R.~Tao, ``{An object fine-grained change detection method based on frequency decoupling interaction for high-resolution remote sensing images},'' \emph{{IEEE} Trans. Geosci. Remote Sens.}, vol.~62, pp. 1--13, 2023.

\bibitem{46F}
S.~Lee, J.~Bae, and H.~Y. Kim, ``{Decompose, adjust, compose: Effective normalization by playing with frequency for domain generalization},'' in \emph{in Proc. {IEEE/CVF} Conf. Comput. Vis. Pattern Recognit. ({CVPR})}, 2023, pp. 11\,776--11\,785.

\bibitem{48F}
B.~Cao, Q.~Wang, P.~Zhu, Q.~Hu, D.~Ren, W.~Zuo, and X.~Gao, ``{Multi-view knowledge ensemble with frequency consistency for cross-domain face translation},'' \emph{{IEEE} Trans. Neural Netw. Learn. Syst.}, vol.~35, no.~7, pp. 9728--9742, 2023.

\bibitem{38f}
J.~Huang, D.~Guan, A.~Xiao, and S.~Lu, ``{Fsdr: Frequency space domain randomization for domain generalization},'' in \emph{in Proc. {IEEE/CVF} Conf. Comput. Vis. Pattern Recognit. ({CVPR})}, 2021, pp. 6891--6902.

\bibitem{frequency3}
D.~Xue, T.~Lei, S.~Yang, Z.~Lv, T.~Liu, Y.~Jin, and A.~K. Nandi, ``{Triple change detection network via joint multifrequency and full-scale swin-transformer for remote sensing images},'' \emph{{IEEE} Trans. Geosci. Remote Sens.}, vol.~61, pp. 1--15, 2023.

\bibitem{WS-Net++}
F.~Xiong, T.~Li, Y.~Yang, J.~Zhou, J.~Lu, and Y.~Qian, ``{Wavelet siamese network with semi-supervised domain adaptation for remote sensing image change detection},'' \emph{{IEEE} Trans. Geosci. Remote Sens.}, 2024.

\bibitem{frequency2}
Y.~Chen, S.~Feng, C.~Zhao, N.~Su, W.~Li, R.~Tao, and J.~Ren, ``{High-resolution remote sensing image change detection based on {Fourier} feature interaction and multi-scale perception},'' \emph{{IEEE} Trans. Geosci. Remote Sens.}, 2024.

\bibitem{yao2023extended}
J.~Yao, B.~Zhang, C.~Li, D.~Hong, and J.~Chanussot, ``{Extended vision transformer ({ExViT}) for land use and land cover classification: A multimodal deep learning framework},'' \emph{{IEEE} Trans. Geosci. Remote Sens.}, vol.~61, pp. 1--15, 2023.

\bibitem{2024hybrid}
C.~Xu, Z.~Ye, L.~Mei, H.~Yu, J.~Liu, Y.~Yalikun, S.~Jin, S.~Liu, W.~Yang, and C.~Lei, ``{Hybrid attention-aware transformer network collaborative multiscale feature alignment for building change detection},'' \emph{{IEEE} Trans. Instrum. Meas.}, vol.~73, pp. 1--14, 2024.

\bibitem{deformable2}
X.~Zhu, H.~Hu, S.~Lin, and J.~Dai, ``{Deformable convnets v2: More deformable, better results},'' in \emph{in Proc. {IEEE/CVF} Conf. Comput. Vis. Pattern Recognit. ({CVPR})}, 2019, pp. 9308--9316.

\bibitem{deformable5}
Z.~Huang, Y.~Wei, X.~Wang, W.~Liu, T.~S. Huang, and H.~Shi, ``{Alignseg: Feature-aligned segmentation networks},'' \emph{{IEEE} Trans. Pattern Anal. Mach. Intell.}, vol.~44, no.~1, pp. 550--557, 2021.

\bibitem{if16.2}
D.~Wen, X.~Huang, F.~Bovolo, J.~Li, X.~Ke, A.~Zhang, and J.~A. Benediktsson, ``{Change detection from very-high-spatial-resolution optical remote sensing images: Methods, applications, and future directions},'' \emph{{IEEE} Geosci. Remote Sens. Mag.}, vol.~9, no.~4, pp. 68--101, 2021.

\bibitem{ResNet}
K.~He, X.~Zhang, S.~Ren, and J.~Sun, ``{Deep residual learning for image recognition},'' in \emph{in Proc. {IEEE} Conf. Comput. Vis. Pattern Recognit. ({CVPR})}, 2016, pp. 770--778.

\bibitem{Res-like1}
M.~Wang, J.~Zhang, G.~Huang, L.~Lu, and F.~Hua, ``{Change detection based on image standardization and improved residual network for single-polarization {SAR} images},'' \emph{{IEEE} J. Sel. Top. Appl. Earth Observ. Remote Sens.}, 2025.

\bibitem{Unet-like}
C.~Wu, L.~Zhang, B.~Du, H.~Chen, J.~Wang, and H.~Zhong, ``{UNet-Like Remote Sensing Change Detection: A review of current models and research directions},'' \emph{{IEEE} Geosci. Remote Sens. Mag.}, 2024.

\bibitem{MSCA}
M.~Liu, Z.~Chai, H.~Deng, and R.~Liu, ``{A {CNN}-transformer network with multiscale context aggregation for fine-grained cropland change detection},'' \emph{{IEEE} J. Sel. Top. Appl. Earth Observ. Remote Sens.}, vol.~15, pp. 4297--4306, 2022.

\bibitem{FC-EE}
R.~C. Daudt, B.~Le~Saux, and A.~Boulch, ``{Fully convolutional siamese networks for change detection},'' in \emph{in Proc. {IEEE} Int. Conf. Image Process. ({ICIP})}.\hskip 1em plus 0.5em minus 0.4em\relax {IEEE}, 2018, pp. 4063--4067.

\bibitem{SPATIAL}
Y.~Liu, C.~Pang, Z.~Zhan, X.~Zhang, and X.~Yang, ``{Building change detection for remote sensing images using a dual-task constrained deep siamese convolutional network model},'' \emph{{IEEE} Geosci. Remote Sens. Lett.}, vol.~18, no.~5, pp. 811--815, 2020.

\bibitem{CHANNEL}
S.~Fang, K.~Li, J.~Shao, and Z.~Li, ``{SNUNet-CD: A densely connected Siamese network for change detection of {VHR} images},'' \emph{{IEEE} Geosci. Remote Sens. Lett.}, vol.~19, pp. 1--5, 2021.

\bibitem{review2}
T.~Bai, L.~Wang, D.~Yin, K.~Sun, Y.~Chen, W.~Li, and D.~Li, ``{Deep learning for change detection in remote sensing: a review},'' \emph{{Geo-Spat. Inf. Sci.}}, vol.~26, no.~3, pp. 262--288, 2022.

\bibitem{Attention}
A.~Vaswani, N.~Shazeer, N.~Parmar, J.~Uszkoreit, L.~Jones, A.~N. Gomez, {\L}.~Kaiser, and I.~Polosukhin, ``{Attention is all you need},'' \emph{{Adv. Neural Inf. Process. Syst.}}, vol.~30, 2017.

\bibitem{VIT}
A.~Dosovitskiy, L.~Beyer, A.~Kolesnikov, D.~Weissenborn, X.~Zhai, T.~Unterthiner, M.~Dehghani, M.~Minderer, G.~Heigold, S.~Gelly \emph{et~al.}, ``{An image is worth 16x16 words: Transformers for image recognition at scale},'' \emph{arXiv preprint arXiv:2010.11929}, 2020.

\bibitem{Changeformer}
W.~G.~C. Bandara and V.~M. Patel, ``{A transformer-based siamese network for change detection},'' in \emph{in Proc. {IEEE} Int. Geosci. Remote Sens. Symp. ({IGARSS})}.\hskip 1em plus 0.5em minus 0.4em\relax {IEEE}, 2022, pp. 207--210.

\bibitem{swinsunet}
C.~Zhang, L.~Wang, S.~Cheng, and Y.~Li, ``{SwinSUNet: Pure transformer network for remote sensing image change detection},'' \emph{{IEEE} Trans. Geosci. Remote Sens.}, vol.~60, pp. 1--13, 2022.

\bibitem{BIT}
H.~Chen, Z.~Qi, and Z.~Shi, ``{Remote sensing image change detection with transformers},'' \emph{{IEEE} Trans. Geosci. Remote Sens.}, vol.~60, pp. 1--14, 2021.

\bibitem{ConFormer-CD}
F.~Yang, M.~Li, W.~Shu, A.~Qin, T.~Song, C.~Gao, and G.-S. Xia, ``{ConvFormer-CD: Hybrid {CNN}-Transformer with Temporal Attention for Detecting Changes in Remote Sensing Imagery},'' \emph{{IEEE} Trans. Geosci. Remote Sens.}, 2025.

\bibitem{CDNeXt}
J.~Wei, K.~Sun, W.~Li, W.~Li, S.~Gao, S.~Miao, Q.~Zhou, and J.~Liu, ``{Robust change detection for remote sensing images based on temporospatial interactive attention module},'' \emph{{Int. J. Appl. Earth Observ. Geoinf.}}, vol. 128, p. 103767, 2024.

\bibitem{GCFormer}
W.~Yu, L.~Zhuo, and J.~Li, ``{GCFormer: Global context-aware transformer for remote sensing image change detection},'' \emph{{IEEE} Trans. Geosci. Remote Sens.}, vol.~62, pp. 1--12, 2024.

\bibitem{Featureinteraction1}
Y.~Wang, W.~Huang, F.~Sun, T.~Xu, Y.~Rong, and J.~Huang, ``{Deep multimodal fusion by channel exchanging},'' \emph{{Adv. Neural Inf. Process. Syst.}}, vol.~33, pp. 4835--4845, 2020.

\bibitem{MFCD}
C.~Zhao, Y.~Tang, S.~Feng, Y.~Fan, W.~Li, R.~Tao, and L.~Zhang, ``{High-resolution remote sensing bitemporal image change detection based on feature interaction and multitask learning},'' \emph{{IEEE} Trans. Geosci. Remote Sens.}, vol.~61, pp. 1--14, 2023.

\bibitem{47F}
Y.~Yang and S.~Soatto, ``{Fda: Fourier domain adaptation for semantic segmentation},'' in \emph{in Proc. {IEEE/CVF} Conf. Comput. Vis. Pattern Recognit. ({CVPR})}, 2020, pp. 4085--4095.

\bibitem{wavelet1}
D.~Quan, H.~Wei, S.~Wang, Y.~Li, J.~Chanussot, Y.~Guo, B.~Hou, and L.~Jiao, ``{Efficient and robust: A cross-modal registration deep wavelet learning method for remote sensing images},'' \emph{{IEEE} J. Sel. Top. Appl. Earth Observ. Remote Sens.}, vol.~16, pp. 4739--4754, 2023.

\bibitem{SENet}
J.~Hu, L.~Shen, and G.~Sun, ``{Squeeze-and-excitation networks},'' in \emph{in Proc. {IEEE} Conf. Comput. Vis. Pattern Recognit. ({CVPR})}, 2018, pp. 7132--7141.

\bibitem{CDD}
M.~A. Lebedev, Y.~V. Vizilter, O.~V. Vygolov, V.~A. Knyaz, and A.~Y. Rubis, ``{Change detection in remote sensing images using conditional adversarial networks},'' \emph{{Int. Arch. Photogramm. Remote Sens. Spat. Inf. Sci.}}, vol.~42, pp. 565--571, 2018.

\bibitem{wang2025enhancing}
M.~Wang, X.~Zhao, and L.~Ren, ``Enhancing cloud-removed regions in multispectral optical images using sar edge features,'' \emph{{IEEE} Geosci. Remote Sens. Lett.}, 2025.

\bibitem{GZ-CD}
D.~Peng, L.~Bruzzone, Y.~Zhang, H.~Guan, H.~Ding, and X.~Huang, ``{SemiCDNet: A semisupervised convolutional neural network for change detection in high resolution remote-sensing images},'' \emph{{IEEE} Trans. Geosci. Remote Sens.}, vol.~59, no.~7, pp. 5891--5906, 2020.

\bibitem{LEVIR-CD}
H.~Chen and Z.~Shi, ``{A spatial-temporal attention-based method and a new dataset for remote sensing image change detection},'' \emph{{Remote Sens.}}, vol.~12, no.~10, p. 1662, 2020.

\end{thebibliography}

\end{document}